\documentclass{article} 
\usepackage{iclr2023_conference,times}

\usepackage{amsmath,amsfonts,bm}

\def\eqref#1{equation~\ref{#1}}

\def\1{\bm{1}}

\DeclareMathAlphabet{\mathsfit}{\encodingdefault}{\sfdefault}{m}{sl}
\SetMathAlphabet{\mathsfit}{bold}{\encodingdefault}{\sfdefault}{bx}{n}

\usepackage{hyperref}
\usepackage{url}
\usepackage{microtype}
\usepackage{breqn}
\usepackage{caption,tabularx,array,booktabs}
\usepackage{graphicx}
\usepackage{caption}
\usepackage{algorithm}
\usepackage{algpseudocode}
\title{Adversary-Aware Partial label learning with Label distillation}
\usepackage{tikz}
\usetikzlibrary{bayesnet}
\usetikzlibrary{arrows}
\usepackage{color}
\usepackage{graphicx}
\usepackage{subcaption}
\usetikzlibrary{backgrounds}
\usepackage{wrapfig}
\usepackage{xcolor,colortbl}
\usepackage{mathtools}

\author{Cheng Chen \\
School of Computer Science, \\
The University of Technology Sydney, \\
Sydney, 100081 Australia \\
\texttt{\{(e-mail:
cheng.chen-16@student.uts.edu.au;)} \\
\And
Yueming Lyu \\
the A*STAR Centre for Frontier AI Research,\\
Singpore 138632 \\
\texttt{\{(e-mail:
\text{Lyu$\_$Yueming@ihpc.a-star.edu.sg};)} \\
\AND
Ivor W.Tsang Fellow, IEEE \\
School of Computer Science, \\
The University of Technology Sydney, \\
Sydney, 100081 Australia \\
The A*STAR Centre for Frontier AI Research,\\
Singpore 138632,\\
\texttt{\{(e-mail:
\text{ivor$\_$tsang@ihpc.a-star.edu.sg};)}}
\iclrfinalcopy 

\begin{document}
\maketitle
\begin{abstract}
To ensure that the data collected from human subjects is entrusted with a secret, rival labels are introduced to conceal the information provided by the participants on purpose. The corresponding learning task can be formulated as a noisy partial-label learning problem. However, conventional partial-label learning (PLL) methods are still vulnerable to the high ratio of noisy partial labels, especially in a large labelling space. To learn a more robust model, we present Adversary-Aware Partial Label Learning and introduce the $\textit{rival}$, a set of noisy labels, to the collection of candidate labels for each instance. By introducing the rival label, the predictive distribution of PLL is factorised such that a handy predictive label is achieved with less uncertainty coming from the transition matrix, assuming the rival generation process is known. Nonetheless, the predictive accuracy is still insufficient to produce an sufficiently accurate positive sample set to leverage the clustering effect of the contrastive loss function. Moreover, the inclusion of rivals also brings an inconsistency issue for the classifier and risk function due to the intractability of the transition matrix. Consequently, an adversarial teacher within momentum (ATM) disambiguation algorithm is proposed to cope with the situation,
allowing us to obtain a provably consistent classifier and risk function. In addition, our method has shown high resiliency to the choice of the label noise transition matrix. Extensive experiments demonstrate that our method achieves promising results on the CIFAR10, CIFAR100 and CUB200 datasets.
\end{abstract}
\section{Introduction}
Deep learning algorithms depend heavily on a large-scale, true annotated training dataset. Nonetheless, the costs of accurately annotating a large volume of true labels to the instances are exorbitant, not to mention the time invested in the labelling procedures. As a result, weakly supervised labels such as partial labels that substitute true labels for learning have proliferated and gained massive popularity in recent years.
Partial-label learning (PLL) is a special weakly-supervised learning problem associated with a set of candidate labels $\vec{Y}$ for each instance, in which only one true latent label $y$ is in existence. Nonetheless, without an appropriately designed learning algorithm, the limitations of the partial label are evident since deep neural networks are still vulnerable to the ambiguous issue rooted in the partial label problem because of noisy labels
\cite{zhou2018brief,patrini2017making,han2018co}.
As a result, there have had many partial label learning works (PLL)\cite{cour2011learning, hullermeier2006learning, feng2019partial,feng2020provably} successfully solved the ambiguity problem where there is a set of candidate labels for each instance, and only a true label exists. Apart from the general partial label, we have also seen a variety of partial label generations evolved, simulating different real-life scenarios. The independently and uniformly drawing is the one have seen the most \cite{pmlr-v119-lv20a,feng2019partial}. The other problem settings include the instance dependent partial label learning, where each partial label set is generated depending on the instance as well as the true label \cite{xu2021instance}. Furthermore, \cite{pmlr-v119-lv20a} has introduced label specific partial label learning, where the uniform flipping probability of similar instances differs from dissimilar group instances. Overall, the learning objective of the previous works is all about disambiguation. More specifically, the goal is to design a classifier training with partial labels, aiming to correctly label the testing dataset,  hoping the classification performance will be as close as the full supervised learning.

On the contrary, there is a lack of discussion on previous works that shed light on the data privacy-enhancing techniques in general partial label learning. The privacy risk is inescapable; thus, privacy-preserving techniques need to be urgently addressed. Recently, we have seen surging data breach cases worldwide. These potential risks posed by the attacker are often overlooked and pose a detrimental threat to society. For instance, it is most likely for the adversary to learn from stolen or leaked partially labelled data for illegal conduct using the previous proposed partial-label learning methods. Subsequently, it has become an inherent privacy concerns in conventional partial label learning. In this paper, the Adversary-Aware partial label learning is proposed to address and mitigate the ramification of the data breach. In a nutshell, we propose an affordable and practical approach to manually corrupt the collected dataset to prevent the adversary from obtaining high-quality, confidential information meanwhile ensure the trustee has full access to the useful information.
However, we have observed that adversary-aware partial label learning possesses some intrinsic learnability issues. 
Firstly, the intractability is raised from the transition matrix. Secondly, the classifier and risk inconsistency problem has been raised.
Hence, we propose an the Adversarial teacher within momentum (ATM)(In section 2.1), adversary-aware loss function \eqref{15}, and a new ambiguity condition \eqref{0} to counter the issues.

Under the adversary-aware partial label problem setting, the rival is added to a candidate set of labels. To achieve that, we extend the original partial label generation \eqref{equation1} by factorisation to add the rival $Y^{\prime}$. Subsequently, we have the adversary-aware partial label generation established as \eqref{eqn:1}. Then, we decompose the second equation of \eqref{eqn:1} into the rival embedded intractable transition matrix term $Q^{*}$ and class instance-dependent transition matrix $T_{y,y^{\prime}}$, which is $\mathrm{P}(Y^{\prime}=y^{\prime} \mid Y=y, X=x)$. In our problem setting, $\bar{T}_{y,y^{\prime}}$, the class instance-independent transition matrix is utilised, which is defined as $\mathrm{P}(Y^{\prime}=y^{\prime} \mid Y=y)$, with the assumption the rival is generated depending only on $Y$ but instance $X$. Under the assumption, the class instance-independent transition matrix is simplified and mathematically identifiable. Since all the instances share the same class instance-independent transition matrix in practice, such encryption is more affordable to implement. The rival variable serves as controllable randomness to enhance privacy against the potential adversary and information leakage. In contrast, the previous methods can not guarantee
the privacy protection property.

However, a fundamental problem has been raised, inclusion of the rival implies an inconsistent classifier according to the adversary-aware label generation equation \eqref{eqn:1}.
Learning a consistent partial label classifier is vital, but in our problem setting, the consistency classifier may not be obtained due to the intractability of $Q^{*}$(details are described in section 1.2). As a consequence,  the Adversarial teacher within momentum (ATM) is proposed, which is designed to identify the term $\mathrm{P}(\vec{Y} \mid Y,Y^{\prime}, X)$ which is denoted as $Q^{*}$. The Moco-style dictionary technique \cite{he2020momentum} and \cite{wang2022pico} have inspired us to explore exploiting the the soft label from instance embedding, leveraging $\bar{T}_{y,y^{\prime}}$ to identify or reduce the uncertainty of the $Q^{*}$ due to the property of informational preservation and tractability. Therefore, a consistent partial label learner is obtained if the uncertainty raised from the transition matrix is reduced greatly. 
Specifically, we transform the inference of label generation in Adversary-Aware PLL as an approximation for the transition matrix $Q^{*}$. Ultimately, a tractable solution to the unbiased estimate of $\mathrm{P}(\vec{Y}\mid Y,Y^{\prime},X)$ can be derived. Lastly, we have rigorously proven that a consistent Adversary-Aware PLL classifier can be obtained if $\mathrm{P}(\vec{Y}\mid Y, Y^{\prime}, X)$ and $\mathrm{P}(Y^{\prime}\mid Y)$ are approximated accurately according to \eqref{eqn:1}. 

In this work, we are mainly focusing on identifying the transition matrix term $\mathrm{P}(\vec{Y}\mid Y, Y^{\prime}, X)$. The rival is generated manually for privacy enhancement. Thus the $\mathrm{P}(Y^{\prime}\mid Y)$ is given by design. Overall, our proposed method has not only solved the ambiguity problem in Adversary-Aware PLL but also addressed the potential risks from the data breach by using a rival as the encryption. Our proposed label generation bears some resemblance to local differential privacy \cite{kairouz2014extremal,warner1965randomized}, which aims to randomise the responses. 
The potential application is to randomise survey responses, a survey technique for improving the reliability of responses to confidential interviews or private questions. Depending on the sophistication of the adversary, our method offers a dynamic mechanism for privacy encryption that is more resilient and flexible to face the potential adversary or privacy risk. By learning from the previous attacks, we can design different levels of protection by adjusting the $\bar{T}$ term. The $\textbf{main contributions}$ of the work are summarized:
\begin{itemize}
\item We propose a novel problem setting named adversary-aware partial label learning.
\item We propose a novel Adversary-Aware loss function and the Adversarial teacher within momentum (ATM) disambiguation algorithm. Our proposed paradigm and loss function can be applied universally to other related partial label learning methods to enhance the privacy protection.
\item A new ambiguity condition (\eqref{0}) for Adversary-Aware Partial Label Learning is derived. Theoretically, we proven that the method is a Classifier-Consistent Risk Estimator.
\end{itemize}
\subsection{Related work}
\textbf{Partial Label Learning (PLL)} trains an instance associated with a candidate set of labels in which the true label is included. 
Many frameworks are designed and proposed to solve the label ambiguity issue in partial label learning. The probabilistic graphical model-based methods\cite{zhang2016partial,wang2020understanding,xu2019partial,lyu2019gm} as well as the clustering-based or unsupervised approaches ~\cite{liu2012conditional} are proposed by leveraging the graph structure and prior information of feature space to do the label disambiguation. The average-based perspective methods ~\cite{hullermeier2006learning,cour2011learning,zhang2016partial} are designed based on the assumption of uniform treatment of all candidates; however, it is vulnerable to the false positive label, leading to misled prediction. Identification perspective-based methods \cite{jin2002learning} tackle disambiguation by treating the true label as a latent variable. The representative perspective approach uses the maximum margin method ~\cite{nguyen2008classification,wang2020online,wang2022pico} to do the label disambiguation. Most recently, self-training perspective methods\cite{feng2019partial,pmlr-v139-wen21a,feng2020provably} have emerged and shown promising performance.
In \textbf{Contrastive Learning} \cite{he2020momentum,oord2018representation}, the augmented input is applied to learns from feature of the unlabeled sample data. The learning objective is to differentiate the similar and dissimilar parts of the input, in turn, maximise the learning of the high-quality representations. CL has been studied in unsupervised representation fashion \cite{chen2020general,he2020momentum}, which treats the same classes as the positive set to boost the performance. The weakly supervised learning has also borrowed the concepts of CL to tackle the partial label problem \cite{wang2022pico}. The CL has also been applied to semi-supervised learning \cite{li2020prototypical}.
\subsection{Adversary-Aware Partial Label Problem Setting}
Given the input space $\mathcal{X} \in \mathbb{R}^{d}$ and label space is defined as $\mathcal{Y}$ = [c] $\coloneqq\in\{1 \cdots c\}$ with the number of $c>2$ classes. Under adversary-aware partial labels, each instance $X \in \mathcal{X}$ has a candidate set of adversary-aware partial labels $\vec{Y} \in \vec{\mathcal{Y}}$. The adversary-aware partial label set has space of $\vec{\mathcal{Y}}:=\{ \vec{y} \mid \vec{y} \subset \mathcal{Y}\}$=$2^{[c]}$, in which there is total $2^{[c]}$ selection of subsets in $[c]$. The objective is to learn a classifier with the adversary-aware partially labelled sample $n$, which was i.i.d drawn from the $\mathcal{\vec{D}}=\{({X}_{1}, \vec{{Y}}_{1}), \ldots,({X}_{n}, \vec{{{Y}}}_{n})\}$, aiming that it is able to assign the true labels for the testing dataset. Given instance and the adversary-aware partial label $\vec{{{Y}}}$ the adversary-aware partial label dataset distribution $\vec{{D}}$ is defined as $(X,\vec{{Y}}) \in \mathcal{X} \times \vec{\mathcal{Y}}$. 
The class instance-independent transition matrix $P(Y^{\prime} \mid Y)$ is denoted as $\bar{T} \in \mathbb{R}^{c \times c}$. $\bar{T}_{y,y^{\prime}}=P(Y^{\prime}=y^{\prime} \mid Y=y)$ where $\bar{T}_{y,y}=0, \forall{y^{\prime},y} \in [c]$. The adversary-aware means the designed paradigm can prevent the adversary from efficiently and reliably inferring certain information from the database without the $\bar{T}$, even if the data was leaked. The rival is the controllable randomness added to the partial label set to enhance privacy.
\subsubsection{Assertion Conditions in Label Generation Set}
The following conditions describe the learning condition for adversary-aware partial label. According to \cite{cour2011learning} there needs to be certain degrees of ambiguity for the partial label learning. Lemma 1 is the new ERM learnability condition which is proposed as follows

\begin{equation}
\label{0}
P_{y^{\prime},\bar{y}}:=\mathrm{P}({y^{\prime},\bar{y}}\in{{\vec{{Y}}}} \mid Y^{\prime}=y^{\prime},\bar{Y}=\bar{y},X=x).
\end{equation}
The $y^{\prime}$ is the rival, and $\bar{y}$ is the false positive label that exists in the partial label set. 
It has to be met to ensure the Adversary-Aware PLL problem is learnable with $y^{\prime}$ $\neq$ $y$ and $\bar{y}$ $\neq$ $y$, these conditions ensure the ERM learnability \cite{liu2014learnability} of the adversary-aware PLL problem if there is small ambiguity degree condition. In our case which is that, $P_{y^{\prime},\bar{y}}<1$.
The $y$ is the true label corresponding to each instance $x$.
And $P_{y}$:= $\mathrm{P}({y}\in{{\vec{{Y}}}} \mid Y=y,X=x)$,
where $P_y=1$ to ensure that the ground truth label is in the partial label set with respect to each instance. 
\subsubsection{Label Generation}
In the previous works of partial label generation procedure, only a candidate of the partial label was generated as such.
\\
\textbf{The Standard Partial Label Generation:}
\begin{equation}
\label{equation1}
\begin{split}
&\sum_{y \in Y} \mathrm{P}(\vec{Y}=\vec{y}, Y=y\mid X=x)= \sum_{y \in Y}\mathrm{P}(\vec{Y}=\vec{y} \mid Y=y, X=x) \mathrm{P}(Y=y \mid X=x).\\
&= \sum_{y \in Y}\mathrm{P}(\vec{Y}=\vec{y} \mid Y=y) \mathrm{P}(Y=y \mid X=x),
\end{split}
\end{equation}
where $\mathrm{P}(\vec{Y}=\vec{y} \mid Y=y, X=x)$ is the label generation for the class instance-dependent partial label and $\mathrm{P}(\vec{Y}=\vec{y} \mid Y=y)$ is the standard partial label learning framework.
Then we present the difference between the general partial labels and the adversary-aware partial label.\\
\textbf{The Adversary-Aware Partial Label Generation:}
\begin{equation}
\resizebox{1\hsize}{!}{
\begin{minipage}{\linewidth}
\label{eqn:1}
\begin{align}
&\sum_{y \in Y} \mathrm{P}(\vec{Y}=\vec{y}\mid X=x)=\sum_{y \in Y}\sum_{y^{\prime} \in Y^{\prime}}{\mathrm{P}(\vec{Y}=\vec{y},  Y=y,Y^{\prime}=y^{\prime}\mid X=x)}\nonumber\\\nonumber
&= \sum_{y \in Y}\sum_{y^{\prime} \in Y^{\prime}}\underbrace{\mathrm{P}(\vec{Y}=\vec{y} \mid Y=y,Y^{\prime}=y^{\prime}, X=x)}_{\textbf{Adversary-Aware transition matrix}}\bar{T}_{y,y^{\prime}}\mathrm{P}(Y=y \mid X=x).\nonumber\nonumber
\end{align}
\end{minipage}}
\end{equation}

In the adversary-aware partial label problem setting, the transition matrix of the adversary-aware partial label is defined as $\mathrm{P}(\vec{Y} \mid Y,Y^{\prime}, X)$ and denoted as $Q^{*} \in \mathbb{R}^{c\times(2^{c}-2)}$. The partial label transition matrix $\mathrm{P}(\vec{Y} \mid Y)$ is denotes as $\bar{Q}\in \mathbb{R}^{c\times(2^{c}-2)}$. Theoretically, if the true label $Y$ of the vector $\vec{Y}$ is unknown given an instance $X$, where $\vec{y} \in {\vec{Y}}$ and there are $ 2^{c}-2$ candidate label sets.The $\epsilon_{x}$ is the instance-dependent rival label noise for each instance where $\epsilon_{x} \in \mathbb{R}^{1\times c}$. The entries of the adversary-aware transition matrix for each instance is defined as follows

\begin{equation}
\label{3}
\sum_{j=1}^{2^c-2}{Q^{*}[:,j]}=
\sum_{j=1}^{2^c-2}([\bar{Q}[:,j]^{T}+\epsilon_{x} ]\bar{T})^{T}=
\sum_{j=1}^{2^c-2}(A[:,j]^{T}\bar{T})^{T},
\end{equation}
where $A[:,j]^{T}=\bar{Q}[:,j]^{T}+\epsilon_{x}$. By formulating the rival as $\boldsymbol{Q^{*}}=\mathrm{P}(\vec{Y}\mid Y^{\prime}, Y ,X)$, which equal to
$\displaystyle\min\{1,A^{(2^{c}-2) \times c }\bar{T}^{c,c}\}$
and $Q^{*}_{i,j} \in[0,1]^{(2^{c}-2) \times c}$, for $\forall_{i,j} \in[c]$. We now have the adversary aware partial label. The conditional distribution of the adversary-aware partial label set $\vec{Y}$ based on \cite{pmlr-v139-wen21a} is derived as belows
\begin{equation}
\label{eqn:5}
\resizebox{1\hsize}{!}{
\begin{minipage}{\linewidth}
\begin{align}
\mathrm{P}(\vec{Y}=\vec{y} \mid Y=y,Y^{\prime}=y^{\prime}, X=x)=\prod_{b^{\prime} \in \vec{y}, b^{\prime} \neq y} p_{b^{\prime}} \cdot \prod_{t^{\prime} \notin \vec{y}}\left(1-p_{t^{\prime}}\right),\nonumber
\end{align}
\end{minipage}}
\end{equation}

where $p_{t^{\prime}}$ and $p_{b^{\prime}}$ are defined as
\begin{equation}
\label{eqn:6}
p_{t^{\prime}} :=\mathrm{P}({t}\in{{\vec{{Y}}}} \mid Y=y,Y^{\prime}=y^{\prime},X=x)<1,
p_{b^{\prime}} :=\mathrm{P}({b}\in{{\vec{{Y}}}} \mid Y=y,Y^{\prime}=y^{\prime},X=x)<1.
\end{equation} 

We summarize the \eqref{eqn:1} as a matrix form in \eqref{eqn:7}.

The inverse problem is to identify a sparse approximation matrix $\boldsymbol{A}$ to use \eqref{ArgMaxEq} to estimate the true posterior probability.
\begin{equation}
\label{eqn:7}
\begin{minipage}{\linewidth}
\begin{align}
&\underbrace{P( \vec{Y} \mid X=x)}_{\textbf{Adversary-aware PLL}}\nonumber= 
\boldsymbol{Q^{*}}\underbrace{P(Y \mid X=x)}_{\textbf{True posterior probability}}\nonumber,
\\
&\boldsymbol{{Q^{*}}}^{-1}\underbrace{P( \vec{Y} \mid X=x)}_{\textbf{Adversary-aware PLL}}\nonumber= 
\underbrace{P(Y \mid X=x)}_{\textbf{True posterior probability}}\nonumber,
\end{align}
\end{minipage}
\end{equation}
\begin{equation}
\label{ArgMaxEq}
\resizebox{0.9\hsize}{!}{
\begin{minipage}{\linewidth}
\begin{align}
&\boldsymbol{\bar{T}}^{-1}\boldsymbol{A}^{-1}  \underbrace{P( \vec{Y} \mid X=x)}_{\textbf{Adversary-aware PLL}}\nonumber \approx \underbrace{P(Y \mid X=x)}_{\textbf{True posterior probability}}\nonumber.
\end{align}\end{minipage}}
\end{equation}
In reality, due to the computational complexity of the transition matrix, it would be a huge burden to estimate $Q^{*}$ accurately for each instance. The $2^{c}-2$ is an extremely large figure and increases exponentially as the label space increase. Therefore, we are no longer required to estimate the true transition matrix $\mathrm{P}(\vec{Y} \mid Y, Y^{\prime}, X)$. Instead, we resort to using instance embedding in the form of a soft label to identify the adversary-aware partial label transition matrix $Q^{*}$. Specifically, we proposed to use a soft pseudo label from the instance embedding (Prototype) to approximate the adversary-aware transition matrix for each instance. The reason is that we can not achieve the true transition matrix $Q^{*}$ directly due to the nature of the practical partial label problem. Therefore, we have used the self-attention prototype learning to approximate the true transition matrix. The detail is described in section 2.1.
Since the Adversary-aware partial label is influenced by the rival label noise, it is challenging to accurately estimate both the class instance-independent transition matrix $\boldsymbol{\bar{T}}$ and the sparse matrix $\boldsymbol{A}$ simultaneously to estimate the true posterior. Considering that the $\boldsymbol{\bar{T}}$ is private and given, it is easier for us just to approximate $\boldsymbol{A}$ to estimate the posterior probability than the adversary. The  \eqref{ArgMaxEq} is implemented as the loss function in \eqref{eqn:loss1}.
\subsection{Positive Sample Set}
The construction of a positive sample is used for contrastive learning to identify the transition matrix $P(\vec{Y}\mid Y^{\prime}, Y, X)$ via the label disambiguation.
Nonetheless, the performance of the contrastive learning erodes drastically due to the introduced rival, which is manifest in the poorly constructed positive sample set, resulting in the degenerated classification performance (See Figure 2). Subsequently, the adversary-aware loss function is proposed in conjunction the contrastive learning to prevent classification performance degeneration.
To start with, we define $L_{2}$ norm embedding of $u$ and $k$ as the query and key latent feature from the feature extraction network $\boldsymbol{f}_{\Theta}$ and key neural network $f_{\Theta}^{\prime}$ respectively. Correspondingly, we have the output $\boldsymbol{u}\in R^{1 \times d}$ where $\boldsymbol{u}_{i}=f_{\Theta}(\operatorname{Aug}_{q}(x))$ and $\boldsymbol{z} \in R^{1 \times d}$ where 
$\boldsymbol{z}_i$=$f_{\Theta}^{\prime}(\operatorname{Aug}_{k}(\boldsymbol{x}_{i}))$. 
The construction of a positive sample set is shown as follows. In each mini-batch, we have $\vec{D}_{b}$ where $\vec{D}_{b} \in \vec{D}$. The $f(x_{i})$ is the function of a neural network with a projection head of 128 feature dimensionality. The outputs of $D_{q}$ and $D_{k}$ are defined as follows,
\begin{align}
&D_{q}=\{\boldsymbol{u}_{i}=\mathbf{ }f\left(\operatorname{Aug}_{q}\left(\boldsymbol{x}_{i}\right)\right) \mid \boldsymbol{x}_{i} \in \vec{D_{b}}\},\\
&D_{k}=\{\boldsymbol{z}_{i}=\mathbf{ }f^{\prime}\left(\operatorname{Aug}_{k}\left(\boldsymbol{x}_{i}\right)\right) \mid \boldsymbol{x}_{i} \in \vec{D_{b}}\},
\end{align}
where $\bar{S}(\boldsymbol{x})$ is the sample set excluding the query set ${q}$ and is defined as $\bar{S}(\boldsymbol{x})=\bar{C}\backslash\{\boldsymbol{q}\}$, in which $\bar{\mathcal{C}}=D_{q} \cup D_{k} \cup \text { queue }$. The $D_{q}$ and $D_{k}$ are vectorial embedding with respect to the query and key views given the current mini-batch. The queue size is determined accordingly depending on the input. The instances from the current mini-batch with the prediction label $\bar{y}^{\prime}$ equal to $(\hat{y}_{i}=c)$ from the $\mathcal{\bar{S}}(x)$. is chosen to be the positive sample set. Ultimately, the $N(\boldsymbol{x})$ is acquired, and it is denoted as
\begin{align}
N_{+}{\left(\boldsymbol{x}_{i}\right)}=\left\{\boldsymbol{z}^{\prime} \mid \boldsymbol{z}^{\prime} \in {\mathcal{\bar{S}}}\left(\boldsymbol{x}_{i}\right), \bar{y}^{\prime}=(\hat{y}_{i}=c)\right\}.
\end{align}
The $N_{+}(x)$ is the positive sample set. The construction of sufficiently accurate positive sample set $N_{+}(x)$ is vital as it underpins the clustering effect of the latent embedding in the contrastive learning procedure. The quality of the clustering effect relies on the precision of prototype $v_{j}$ corresponding to $j\in\{1,...,C\}$. Our method helps maintain the precision of prototypes using the $\bar{T}$ to render better label disambiguation module performance for contrastive learning when introduced the rival. 
where the query embedding $u$ multiplies the key embedding $z$ and then divides with the remaining pool $\bar{C}$. 
Overall, the $S_{+}(x)$ is used to facilitate the representation learning of the contrastive learning and the self-attention prototype learning to do the label disambiguation or a more accurate pseudo-labelling procedure.
Our proposed loss ensures the prototype and contrastive learning are working systematically and benefit mutually when the rival is introduced. 
The pseudo label generation is according to  \eqref{eqn:13}. We have followed \cite{wang2022pico} for the positive sample selection.
\section{Methodology}
\begin{figure*}
\centering
\captionsetup{font=small}
{\includegraphics[scale=0.48]{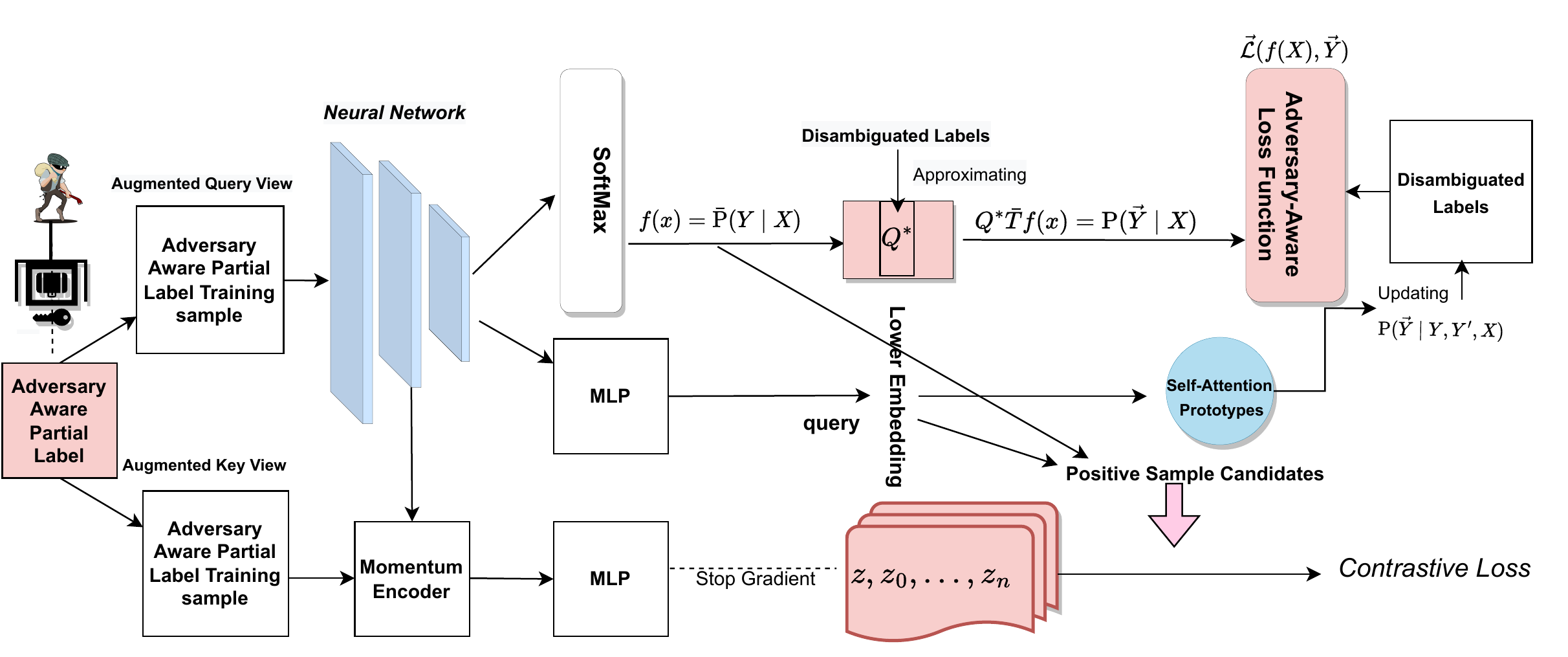}}\hspace{1em}%
\caption{\footnotesize An overview of the proposed method. General partial label can be disclosed to adversary. 
The initial training is about positive sample selection. Moreover, we have assumed $\bar{T}$ is given. 
}
\label{est_error}
\end{figure*}
The main task of partial label learning is label disambiguation, which targets identifying the true label among candidate label sets.
Thus, we present an adversarial teacher within momentum (ATM). 
The \eqref{eqn:loss1} is developed to do the debiasing from the prediction of $f(x)$ given the adversary-aware partial label via the class instance dependent transition matrix $\bar{T}+I$. The unbiased prediction induces the identification of a more accurate positive sample set which allows Equation 18 to leverage the high-quality presentation power of a positive sample set to improve the classification performance.
\subsection{ Pseudo Label Learners via Adversarial Teacher within Momentum (ATM)}
Unlike \cite{wang2022pico}, we present an adversarial teacher strategy with momentum update (ATM) to guide the learning of pseudo labels using Equation 17. Just like a tough teacher who teaches the subject using harsh contents to test students' understanding of the subject. In our case, the rival is like the subject which is purposely generated by us, at the same time Equation 17 is introduced to check the understanding of the student (classifier) given the scope of testing content which is the $\bar{T}$.
Specifically, the spherically margin between prototype vector $\boldsymbol{v}_i \in \mathbb{S}^{d-1}$ and prototype vector $\boldsymbol{v}_j \in \mathbb{S}^{d-1}$ is defined as
\begin{align}
    m_{i j} =  \exp{(-\boldsymbol{v}_i^\top \boldsymbol{v}_j)}.
\end{align}
For prototype $\boldsymbol{v}_i$, we define the normalized margin between  $\boldsymbol{v}_i$ and  $\boldsymbol{v}_j$ as 
\begin{align}
    \bar{m}_{i j} = \frac{ \exp{(-\boldsymbol{v}_i^\top \boldsymbol{v}_j)}}{ \sum_ {j \ne i} {  \exp{(-\boldsymbol{v}_i^\top \boldsymbol{v}_j)}}}.
\end{align}
For each $\boldsymbol{v}_i,i\in\{1, \cdots, K\}$, we perform momentum updating with the normalized margin between $\boldsymbol{v}_j$ and $\boldsymbol{v}_i$ for all $j \ne i$ as an regularization. The resulted new update rule is given as 
\begin{align}
    \label{10}
    \boldsymbol{v}_i^{t+1} = \sqrt{1-\alpha^2} \boldsymbol{v}_i^{t} +  \alpha  \frac{\boldsymbol{g}}{\| \boldsymbol{g}\|_2},
\end{align}
where the gradient $\boldsymbol{g}$ is given as 
\begin{align}
    \label{11}
    \boldsymbol{g} = \boldsymbol{u} - \beta \sum_{j \ne i} { \bar{m}_{ij}^t \boldsymbol{v}_j^t },
\end{align}
where $\boldsymbol{u}$ is the query embedding whose prediction is class $i$, $\bar{m}_{ij}^t$ is the normalized margin between prototype vectors at step $t$ (i.e., $\boldsymbol{v}_j^t, j \ne i$). The $v_{c}$ is the prototype corresponding to each class.
\begin{align}
\label{eqn:13}
\boldsymbol{\bar{q}}=\phi \boldsymbol{\bar{q}}+(1-\phi) \boldsymbol{v}, \quad v_{c}= \begin{cases}1 & \text { if } c=\arg \max _{j \in Y} \boldsymbol{u}^{\top} \boldsymbol{v} \\ 0 & \text { otherwise ,}\end{cases}.
\end{align}
where $\bar{q}$ is the target prediction and subsequently used in the \eqref{eqn:loss1}. It was initialised as the uniform probability $\boldsymbol{\bar{q}}=\frac{1}{|c|} \mathbb{1}$ and updated accordingly to the 
\eqref{eqn:13}. The $\phi$ is the hyper-parameter controlling for the updating of  $\boldsymbol{\bar{q}}$. 
\subsection{Adversary Aware Loss Function.} 
The goal is to build a risk consistent loss function, hoping it can achieve the same generalization error as the supervised classification risk $R(f)$ with the same classifier $f$.
To train the classifier, we minimize the following modified loss function estimator by leveraging the updated pseudo label from the Adversarial teacher within momentum (ATM) distillation method and transition plus identity matrix, $I_{i,j} \in[0,1]^{c \times c}$, $I_{i,i}=1$, for $\forall_{i=j} \in[c]$, $I_{i,j}=0$, for $\forall_{i\neq j} \in[c]$:
where $f\left(\boldsymbol{X}\right)\in\mathbb{R}^{\lvert c \lvert}$,
\begin{equation}
\label{eqn:loss1}
\resizebox{1\hsize}{!}{
\begin{minipage}{\linewidth}
\begin{align}
\vec{\mathcal{L}}(f(X), \vec{Y}) =-\sum_{i=1}^{c} ({{\bar{{q_{i}}}}})\log \left(((\mathbf{\bar{T}+I}) {f(X))_{i}}\right)\nonumber.
\end{align}
\end{minipage}}
\end{equation}
The proof for the modified loss function is shown in the appendix lemma 4.
In our case, given sufficiently accurate positive sample set of the contrastive learning is utilised to incorporate with \eqref{eqn:loss1} to identify the transition matrix of the adversary-aware partial label. The contrastive loss is defined as follows
\begin{equation}
\label{eqn:14}
\resizebox{1\hsize}{!}{
\begin{minipage}{\linewidth}
\begin{align}
&\mathcal{L}_{\mathrm{}}(f(x ), \tau, C)\nonumber= \frac{1}{ |D_{q}|} \sum_{\boldsymbol{u}  \in D_{q}} \{-\frac{1}{ N_{+}(x)}\sum_{\boldsymbol{z_{+}} \in N_{+}(x)}\log \frac{\exp(\boldsymbol{u}^{\top} \boldsymbol{z}/\tau)}{\sum_{\boldsymbol{z}^{\prime} \in \bar{C}(\boldsymbol{x})}\exp(\boldsymbol{u}^{\top} \boldsymbol{z} / \tau)}\}\nonumber.
\end{align}
\end{minipage}}
\end{equation}

Finally, we have the Adversary-Aware Loss expressed as
\begin{equation}
\begin{aligned}
\label{15}
&\text{\textbf{Adversary-Aware Loss}}=\lambda\mathcal{L}_{\mathrm{}}(f(x_{i}), \tau, C)+\vec{\mathcal{L}}(f(X), \vec{Y}).
\end{aligned}
\end{equation}

There are two terms of the proposed loss function (\eqref{15}), which are the \eqref{eqn:loss1} and  \eqref{eqn:14} correspondingly. 
\eqref{eqn:loss1} is developed to lessen prediction errors from $f(x)$ given the adversary-aware partial label. The debiasing is achieved via the class instance dependent transition matrix $\bar{T}+I$ by down-weighting the false prediction. The unbiased prediction induces the identification of a more accurate positive sample set. \eqref{eqn:14} is the contrastive loss. It leverages the high-quality representation power of positive sample set to improve the classification performance further.
\section{Theoretical Analysis}
The section introduces the concepts of classifier consistency and risk consistency ~\cite{xia2019anchor} ~\cite{zhang2004statistical}, which are crucial in weakly supervised learning. Risk consistency is achieved if the risk function of weak supervised learning is the same as the risk of fully supervised learning with the same hypothesis. The risk consistency implies classifier consistency, meaning classifier trained with partial labels is consistent as the optimal classifier of the fully supervised learning.

\textbf{Classifier-Consistent Risk Estimator} 
\textbf{Learning with True labels.} Lets denote $f(X)=\left(g_{1}(x), \ldots, g_{K}(x)\right)$ as the classifier, in which $g_{c}(x)$ is the classifier for label $c \in[K]$. The prediction of the classifier $f_{c}(x)$ is $P(Y=c\mid x)$. We want to obtain a classifier $f(X)$ =$\arg \max_{i\in[K]}{g_{i}(x)}$. The loss function is to measure the loss given classifier $f(X)$. To this end, the true risk can be denoted as
\begin{equation}
\begin{aligned}
R(f)=\mathbb{E}_{(X, {Y})} [\mathcal{L}\left(f\left({X}\right), {Y}\right)].
\end{aligned}
\end{equation}
The ultimate goal is to learn the optimal classifier $f^{*}$=$\arg \min _{f \in \mathcal{F}} R(f)$ for all loss functions, for instance to enable the empirical risk $\bar{R}_{pn}(f)$ to be converged to true risk $R(h).$ To obtain the optimal classifier, we need to prove that the modified loss function is risk consistent as if it can converge to the true loss function.
\\
\\
\textbf{Learning with adversary-aware Partial Label.}
An input $X \in$ $\mathcal{X}$ has a candidate set of $\vec{Y} \in \mathcal{\vec{Y}}$ but a only true label $Y \in \mathcal{\vec{Y}}$. 
Given the adversary-aware partial label $\vec{{Y}} \in \vec{\mathcal{\mathcal{{Y}}}}$ and instance $X \in \mathcal{X}$ that the objective of the loss function is denoted as
\begin{equation}
\begin{aligned}
{\hat{R}}(f)=\mathbb{E}_{(X, \vec{{Y}})} \vec{\mathcal{L}} \left(f\left({X}\right), \vec{{Y}}\right).
\end{aligned}
\end{equation}
Since the true adversary-aware partial label distribution $\bar{\mathcal{D}}$ is unknown, our goal is approximate the optimal classifier with sample distribution $\bar{D}_{pn}$ by minimising the empirical risk function, namely
\begin{equation}
\begin{aligned}
\hat{R}_{pn}(f)=\frac{1}{n} \sum_{i=1}^{n} \vec{\mathcal{L}}\left(f\left(\boldsymbol{x}_{i}\right), \vec{{y}}_{i}\right).
\end{aligned}
\end{equation}

\vspace{0.1cm}

\noindent $\textbf{Assumption 1.}$
According to \cite{yu2018learning} that the minimization of the expected risk $R(f)$ given clean true population implies that the optimal classifier is able to do the mapping of $f_{i}^{*}(X)=P(Y=i \mid X)$, $\forall i \in[c].$
Under the assumption 1, we are able to draw conclusion that $\hat{f}^{*}=f^{*}$ applying the theorem 2 in the following.
\subsubsubsection{Theorem 1}
\noindent 
$\textbf{Theorem 1.}$ \textit{Assume that the Adversary-Aware matrix $\star{T}_{y,y^{\prime}}$ is fully ranked and the Assumption $1$ is met, the the minimizer of $\hat{f}^{*}$ of $\hat{R}(f)$ will be converged to $f^{*}$ of $R(f)$, meaning $\hat{f}^{*}=f^{*}$}.
\noindent $\textbf{Remark.}$
If the $Q^{*}$ and $T_{y,y^{\prime}}$ is estimated correctly the empirical risk of the designed algorithm trained with adversary-aware partial label will converge to the expected risk of the optimal classifier trained with the true label. If the number of sample is reaching infinitely large that given the adversary-aware partial labels, $\hat{f}_{n}$ is going to converged to $\hat{f}^{*}$ theoretically. Subsequently, $\hat{f}_{n}$ will converge to the optimal classifier $f^{*}$ as claimed in the theorem 1.
With the new generation procedure, the loss function risk consistency theorems are introduced. 
\textbf{Theorem 2}. \textit{The adversary-aware loss function proposed is risk consistent estimator if it can asymptotically converge to the expected risk given sufficiently good approximate of $\bar{Q}$ and the adversary-aware matrix. The proof is in appendix lemma 4.}
\begin{align}
\mathcal{L}(y,f(x)) &= \sum_{\vec{y} \in \vec{\mathcal{Y}} y}   \sum_{{y}=1}^{C}\sum_{y^{\prime}\in Y^{\prime}}  \Large( \mathrm{P}(Y=y \mid X=x) \nonumber \\
&= \prod_{b^{\prime} \in \vec{y}} p_{b^{\prime}} \cdot \prod_{t^{\prime} \notin \vec{y}}\left(1-p_{t^{\prime}}\right)\bar{T}_{y,y^{\prime}}\vec{\mathcal{L}}(\vec{y}, f(x)) \Large) \nonumber\\
&=\vec{\mathcal{L}}(\vec{y}, f(x)).
\end{align}

\subsection{Generalisation error}
\textit{Define $\hat{R}$ and ${\hat{R}_{pn}}$ as the true risk the empirical risk respectively given the adversary-aware partial label dataset. The empirical loss classifier is obtained as $\hat{f}_{pn}=\arg \min_{f \in\mathcal{F}}\hat{R}_{pn}(f)$.  Suppose a set of real hypothesis  $\mathcal{F}_{\vec{y}_k}$ with $f_i(X) \in \mathcal{F}, \forall{i}\in [c]$ . Also, assume it's loss function $\vec{\mathcal{L}}(\boldsymbol{f}(X), \vec{Y})$ is ${L}$-Lipschitz continuous with respect to $f(X)$ for all $\vec{y}_{k} \in \vec{\mathcal{Y}}$ and upper-bounded by $M$,
$\text { i.e., } M=\sup _{x \in \mathcal{X}, f \in \mathcal{F}, y_{k} \in \vec{Y}} \vec{\mathcal{L}}\left(f(x), {\vec{y}_{k}}\right)$. The expected Rademacher complexity of $\mathcal{F}{_{k}}$ is denoted as $\Re_{n}(\mathcal{F}_{\vec{y}_k})$}\cite{bartlett2002rademacher}

\textbf{Theorem 3}. \textit{For any $\delta>0$, with probability at least $1-\delta$},
\begin{equation}
\begin{aligned}
    \hat{R}\left(\hat{f}_{pn}\right)-\hat{R}\left({\hat{f}}^{\star}\right)
    & \leq 4 \sqrt{2} L \sum_{k=1}^{c} \Re_{n}\left(\mathcal{F}_{\vec{y}_k}\right)+M \sqrt{\frac{\log \frac{2}{\delta}}{2 n}}.
\end{aligned}   
\end{equation}
As the number of samples reaches to infinity $n\rightarrow \infty, \Re_{n}\left(\mathcal{F}_{\vec{y}_k}\right) \rightarrow 0$ with a bounded norm. Subsequently, $\bar{R}(\hat{f})\rightarrow \bar{R}\left(\hat{f}^{\star}\right)$ as the number of training data reach to infinitely large. The proof is given in Appendix Theorem 3.
\vspace{-1em}
\section{Experiments}
\begin{table*}[h]
\label{Mtable}
\captionsetup{font=small}
\scriptsize\caption {\small
Benchmark datasets for accuracy comparisons. Superior results are indicated in bold. Our proposed methods have shown comparable results to fully supervised learning and outperform previous methods in a more challenging learning scenario, such as the partial rate at 0.5(CIFAR10) and 0.1(CIFAR100, CUB200). The hyper-parameter $\alpha$ is set to 0.1 for our method. (The symbol \textbf{$\ast$} indicates Adversary-Aware partial label dataset).}
\footnotesize
\begin{center}
\begin{tabular}{c|c|c |c |c}\hline
\toprule
Dataset & \footnotesize Method & \footnotesize $q= 0.01$  & \footnotesize $q= 0.05$  & \footnotesize  $q= 0.1$  \\
\midrule
&\footnotesize\text{(ATM)(Without T)(Our)} &  \textbf{73.43}  $\pm{0.11}$ & 72.63 $\pm {0.27} $& \textbf{72.35} $\pm{0.22} $\\
\text{CIFAR100}& \text{PiCO} & 73.28  $\pm{0.24} $   &\textbf{72.90}  $\pm{0.27}$  & 71.77 $\pm{0.14} $ \\
& \text{LWS} & 65.78 $\pm{0.02}$  &59.56 $\pm{0.33}$& 53.53  $\pm{0.08}$\\
& \footnotesize\text{PRODEN} & 62.60 $\pm{0.02} $& 60.73 $\pm{0.03}$& 56.80 $\pm{0.29}$\\
&\footnotesize\text{Full Supervised} &  & 73.56  $\pm {0.10} $ &\\
\midrule
 Dataset & \footnotesize Method & \footnotesize $q^{*}= 0.03\pm{0.02}$  & \footnotesize $q^{*}= 0.05\pm{0.02}$  & \footnotesize  $q^{*}= 0.1\pm{0.02}$  \\
 \midrule
&\footnotesize\text{(ATM)(Our)}$^{\ast}$ &73.36  $\pm {0.32} $ & 72.76 $\pm {0.14} $&  $\textbf{54.09} $  $\pm \mathbf{1.88} $ \\
\footnotesize\text{CIFAR100}& \footnotesize\text{PiCO}$^{\ast}$&72.87
 $\pm{0.26} $ &72.53 $\pm{0.37} $&48.03 $\pm{3.32} $\\
& \footnotesize\text{LWS}$^{\ast}$ & 46.8 $\pm{0.06}$ & 24.82 $\pm{0.17} $ & 4.53 $\pm{0.47} $\\
& \footnotesize\text{PRODEN} $^{\ast}$ &  59.33 $\pm{0.48}$ & 41.20 $\pm{0.27}$ & 13.44$\pm{0.41}$\\
\midrule
\end{tabular}
\end{center}
\label{table:performance:tgif}
\footnotesize
\begin{center}
\resizebox{0.9\linewidth}{!}
{{\small
\begin{tabular}{c|c|c|c|c}
\hline
\toprule
\footnotesize Dataset & \footnotesize Method & \footnotesize $q= 0.01$ & \footnotesize $q= 0.05$ & \footnotesize $q= 0.1$ \\
\midrule
\footnotesize CUB200 & \footnotesize {(ATM) (Without T)(Our)} & \textbf{74.43}$\pm$\textbf{0.876} & \textbf{72.30}$\pm$\textbf{0.521} & \textbf{66.87}$\pm$\textbf{0.98} \\
& \footnotesize PiCO & 74.11$\pm${0.37} & 71.75$\pm${0.56} & 66.12$\pm${0.99} \\
& \footnotesize LWS & 73.74$\pm${0.23} & 39.74$\pm${0.47} & 12.30$\pm${0.77} \\
& \footnotesize PRODEN & 72.34$\pm${0.04} & 62.56$\pm${0.10} & 35.89$\pm${0.05} \\
& \footnotesize Full Supervised & & 76.02$\pm${0.19} & \\
\midrule
\footnotesize Dataset & \footnotesize Method & \footnotesize $q^{*}=\{0.03\pm{0.02}\}$ & \footnotesize $q^{*}=\{0.05\pm{0.02}\}$ & \footnotesize $q^{*}=\{0.1\pm{0.02}\}$ \\
\midrule
& \footnotesize {(ATM) (Our)}$^{\ast}$ & \textbf{72.22}$\pm${1.36} & \textbf{72.43}$\pm${0.86} & \textbf{56.26}$\pm${0.70} \\
\footnotesize CUB200 & \footnotesize PiCO$^{\ast}$ & 71.85$\pm${0.53} & 71.15$\pm${0.41} & 50.31$\pm${1.01} \\
& \footnotesize LWS$^{\ast}$ & 9.6$\pm${0.62} & 4.02$\pm${0.03} & 1.44$\pm${0.06} \\
& \footnotesize PRODEN$^{\ast}$ & 18.71$\pm${0.45} & 17.63$\pm${0.89} & 17.99$\pm${0.62} \\
\midrule
\end{tabular}}}
\end{center}
\label{table:performance:tgif}
\label{table:full}
\footnotesize
\begin{center}
\resizebox{0.9\linewidth}{!}
{{\footnotesize 
\begin{tabular}{c|c|c |c |c}\hline
\toprule
Dataset & \footnotesize Method & \footnotesize $q= 0.1$  & \footnotesize $q= 0.3$  & \footnotesize  $q= 0.5$  \\
\midrule
& \footnotesize\text{(ATM)(Without T)(Our)} &93.57$\pm{0.16}$ & 93.17$\pm{0.09}$ & 92.22$\pm{0.40}$ \\
\text{CIFAR10}& \footnotesize\text{PiCO} & \textbf{93.74}$\pm{0.24}$ & \textbf{93.25}$\pm{0.32}$ & \textbf{92.46}$\pm{0.38}$ \\
& \footnotesize\text{LWS} & 90.30 $\pm{0.60}$ & 88.99 $\pm{1.43}$ & 86.16 $\pm{0.85}$ \\
& \footnotesize\text{PRODEN} & 90.24$\pm{0.32}$ & 89.38$\pm{0.31}$& 87.78$\pm{0.07}$ \\
& \footnotesize\text{Full Supervised} & & 94.91$\pm{0.07}$& \\
\midrule
Dataset & \footnotesize Method & \footnotesize $q^{*}= 0.1\pm{0.02}$ & \footnotesize $q^{*}= 0.3\pm{0.02}$ & \footnotesize $q^{*}= 0.5\pm{0.02}$ \\
\midrule
& \footnotesize{ (ATM) (Our)} $^{\ast}$ & 93.52 $\pm{0.11}$ & \textbf{92.98}$\pm{0.51}$ & \textbf{89.62}$\pm{0.79}$ \\
\footnotesize\text{CIFAR10}& \footnotesize\text{PiCO}$^{\ast}$ & \textbf{93.64}$\pm{0.24}$ &{92.85}$\pm{0.43}$ & 81.45$\pm{0.57}$ \\
& \footnotesize\text{LWS}$^{\ast}$ & 87.34$\pm{0.87}$ & 39.9$\pm{0.72}$ & 9.89$\pm{0.55}$ \\
& \footnotesize\text{PRODEN}$^{\ast}$ & 88.80$\pm{0.14}$ & 81.88$\pm{0.51}$ & 20.32$\pm{3.43}$\\
\toprule
\end{tabular}}}
\end{center}
\label{table:performance:tgif}
\end{table*}
\textbf{Datasets}  
We evaluate the proposed method on three benchmarks-CIFAR10, CIFAR100 \cite{krizhevsky2009learning}, and fine-grained CUB200 \cite{wah2011caltech} with general partial label and adversary-aware partial label datasets.  

\textbf{Main Empirical Results for CIFAR10.} All the classification accuracy is shown in Table~1.
We have compared classification results on CIFAR-10 with previous works \cite{wang2022pico,pmlr-v119-lv20a,pmlr-v139-wen21a} using the Adversarial teacher within momentum (ATM). The method has shown consistently superior results in all learning scenarios where $q=\{0.3,0.5\}$ for the adversary-aware partial label learning. More specifically, the proposed method achieves $\textbf{8.17}\%$ superior classification performance at a 0.5 partial rate than the previous state of art work \cite{wang2022pico}. Moreover, our proposed method has achieved comparable results at 0.1 and 0.3 partial rates. 
The experiments for CIFAR-10 have been repeated four times with four random seeds.
\textbf{Main Empirical Results for CUB200 and CIFAR100.} The proposed method has shown superior results for the Adversary-Aware Partial Label, especially in more challenging learning tasks like the 0.1 partial rate of the dataset cub200 and CIFAR100, respectively. On the cub200 dataset, we have shown \textbf{5.95\%} improvement at partial rates 0.1 and 1.281\% and 0.37\% where the partial rate is at 0.05 and 0.03. On the CIFAR100 dataset, the method has shown \textbf{6.06\%} and 0.4181\%, 0.5414\% higher classification margin at partial rate 0.1, 0.05 and 0.03.The experiments have been repeated five times with five random seeds. 
\subsection{Ablation Study}
Figure 2 shows the experimental result comparisons for CUB200 between the adversary-aware loss function and previous loss function before and after the momentum updating.
Given \eqref{eqn:loss1}, the uncertainty of the transition matrix $\bar{Q}$ is reduced, leading to a good initialisation for the positive set selection, which is a warm start and  plays a vital role in improving the performance of contrastive learning. After we have a good set of positive samples, the prototype's accuracy is enhanced. Subsequently, leveraging the clustering effect and the high-quality representation power of the positive sample set of  contrastive loss function to improve the classification performance. 
\begin{figure}[H]
\centering
\captionsetup{font=scriptsize}
\subcaptionbox{\label{fig1:a}}{\includegraphics[width=0.3\textwidth]{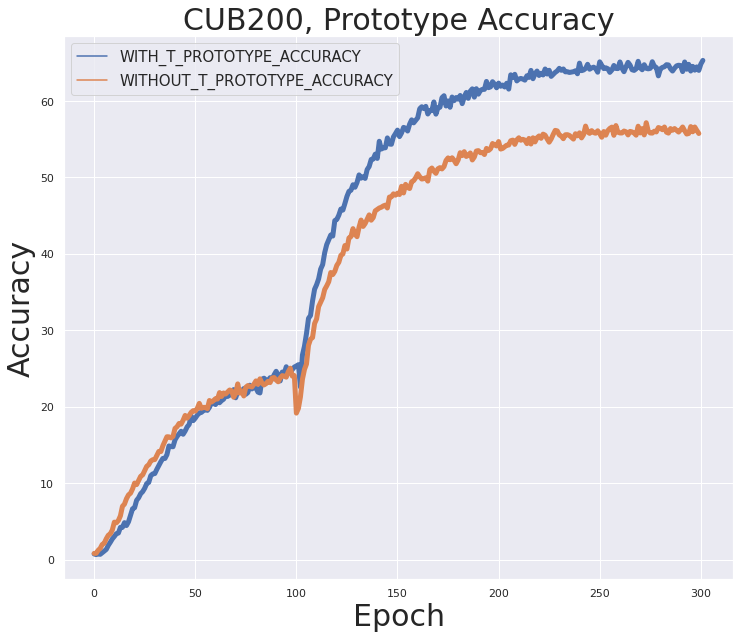}}\hspace{0.2em}%
\subcaptionbox{\label{fig1:b}}{\includegraphics[width=0.3\textwidth]{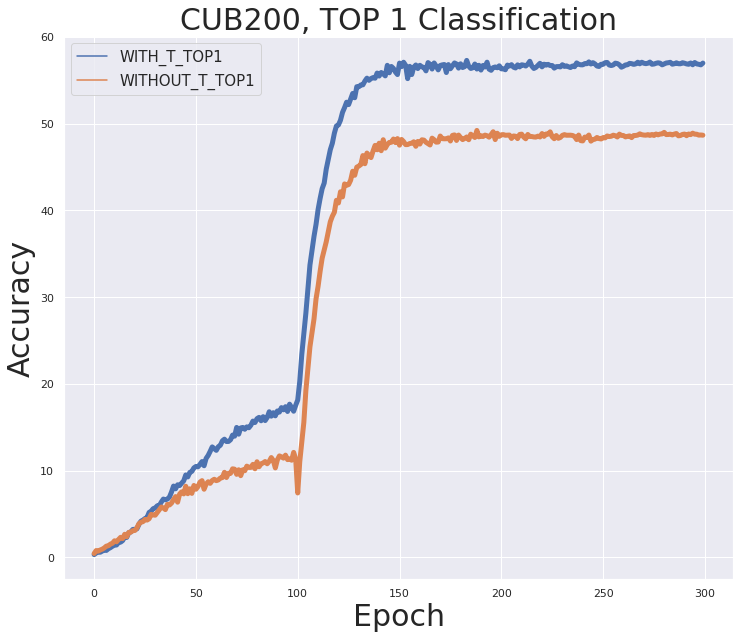}}\hspace{0.2em}%
\caption{The Top1 and Prototype Accuracy of the Proposed Method and the Method in \cite{wang2022pico} on CUB200 Adversary-Aware Loss Comparison. 
}
\label{est_error}
\end{figure}
\section{Conclusion and Future works}
This paper introduces a novel Adversary-Aware partial label learning problem. The new problem setting has taken local data privacy protection into account. Specifically, we have added the rival to the partial label candidate set as encryption for the dataset. Nonetheless, the generation process has made the intractable transition matrix even more complicated, leading to an inconsistency issue. Therefore, the novel adversary-aware loss function and the self-attention prototype are proposed. The method is proven to be a provable classifier and has shown superior performance. Future work will use variational inference methods to approximate the intractable transition matrix.
\newpage
\appendix

\bibliography{iclr2023_conference}
\bibliographystyle{iclr2023_conference}
\section{Appendix}
This file includes supplementary for all proofs and additional experiment details. 
The proofs for Theorem 1, Theorem 2, Lemma 4, Theorem 3, Lemma 4a, Lemma 4b and Additional Experiment are presented sequentially. 
\subsection{\textbf{The Proof for Theorem 1}}
Our goal is to find the optimal classifier, namely
\begin{equation}
\label{eqn:5}
\resizebox{1\hsize}{!}{
\begin{minipage}{\linewidth}
\begin{align}
    &h_{i}^{*}(X)=\mathrm{P}(Y=y\mid X=x) \forall i \in[c]\nonumber
\end{align}
\end{minipage}
}
\end{equation}
We can obtained the optimal classifier with modified loss function (equation 17) and assumption 1 when learning from examples with adversary-aware partial labels. The transition matrix of the adversary-aware partial label is defined as $\mathrm{P}(\vec{Y} \mid Y,Y^{\prime}, X)$ and denoted as $Q^{*} \in \mathbb{R}^{c\times(2^{c}-2)}$. The partial label transition matrix $\mathrm{P}(\vec{Y} \mid Y)$ is denotes as $\bar{Q}\in \mathbb{R}^{c\times(2^{c}-2)}$. Theoretically, if the true label $Y$ of the vector $\vec{Y}$ is unknown given an instance $X$, where $\vec{y} \in {\vec{Y}}$ and there are $ 2^{c}-2$ candidate label sets.The $\epsilon_{x}$ is the instance-dependent rival label noise for each instance where $\epsilon_{x} \in \mathbb{R}^{1\times c}$. 
The class instance-dependent transition matrix is defined as $\bar{T}_{yy^{\prime}} \in[0,1]^{C \times C}$, in which $\bar{T}_{yy^{\prime}}$=  $\mathrm{P}(Y^{\prime}=y^{\prime} \mid Y=y)$ and we assume $\bar{T}_{yy}=0$, for $\forall_{yy^{\prime}} \in[c]$,
The inverse problem is to identify a sparse approximation matrix $\boldsymbol{A}$ given $\bar{T}$ to estimate the true posterior probability. 
\begin{align}
    &\underbrace{P( \vec{Y} \mid X)}_{\textbf{Adversary-aware PLL}}\nonumber=(\boldsymbol{[\bar{Q}^{\text{T}}+\epsilon]\bar{T}})\underbrace{P(Y \mid X)}_{\textbf{True Posterior Probability}},
\end{align}
\begin{align}
    & \boldsymbol{\bar{T}}^{-1}\boldsymbol{A}^{-1}\underbrace{P(\vec{Y} \mid X=x)}_{\textbf{Adversary-aware PLL}}\nonumber\approx \underbrace{P(Y \mid X=x)}_{\textbf{True Posterior Probability}},
\end{align}
which further ensures
\begin{equation}
\label{eqn:7}
\resizebox{1\hsize}{!}{
\begin{minipage}{\linewidth}
\begin{align}
&\underbrace{P(\vec{Y} \mid X)}_{\textbf{Adversary-aware PLL}}=([\bar{Q}^{T}+\epsilon]\bar{T})\underbrace{\mathbf{h}^{*}(X)}_{\textbf{True Posterior Probability}.}\nonumber
\end{align}
\end{minipage}}
\end{equation}
where $Q^{*}=([\bar{Q}^{T}+\epsilon]\bar{T})^{T}$.
If the transition matrix $\mathbf{\bar{T}}$ is full rank and $Q^{*}$ is identified, then we can define the optimal classifier ${h}^{*}(X)=\mathrm{P}(Y=y \mid X=x)$, which guarantees $\hat{f}^{*}=f^{*}$.
The proof is completed.
\newpage
\subsection{The Proof for Theorem 2}
for any $x \in \mathcal{X}$, there holds
\begin{equation}
\resizebox{1\hsize}{!}{
\begin{minipage}{\linewidth}
\begin{align}
& \hat{\mathcal{R}}(\vec{\mathcal{L}}, f(X)) \nonumber\\\nonumber
=& \mathbb{E}_{{\vec{Y}} \mid X}[\vec{\mathcal{L}}(\vec{Y}, f(x)) \mid X=x] \\\nonumber
=& \sum_{\vec{y} \in 2^{[C]}} \vec{\mathcal{L}}(\vec{y}, f(x)) \mathrm{P}(\vec{Y}=\vec{y} \mid X=x) \\\nonumber
=& \sum_{\vec{y} \in 2^{[C]}} \vec{\mathcal{L}}(\vec{y}, f(x)) \sum_{y \in Y}  \mathrm{P}(\vec{Y}=\vec{y}, Y=y \mid X=x) \\\nonumber
=& \sum_{\vec{y} \in 2^{[C]}} \vec{\mathcal{L}}(\vec{y}, f(x)) \sum_{y \in Y} \sum_{y^{\prime}\in Y^{\prime}} \mathrm{P}(\vec{Y}=\vec{y}, Y=y,Y^{\prime}=y^{\prime} \mid X=x) \\\nonumber
=& \sum_{\vec{y} \in 2^{[C]}} \vec{\mathcal{L}}(\vec{y}, f(x)) \\\nonumber
&  (\sum_{y \in Y}\sum_{y^{\prime}\in Y^{\prime}}  \mathrm{P}(\vec{Y}=\vec{y} \mid Y=y,Y^{\prime}=y^{\prime}, X=x) \mathrm{P}(Y^{\prime}=y^{\prime} \mid Y=y, X=x)  \mathrm{P}(Y=y \mid X=x)) \\\nonumber
=& \sum_{{y}=1}^{C} \mathrm{P}(Y=y \mid X=x) \\\nonumber
& (\sum_{\vec{y} \in 2^{[C]}}\sum_{y^{\prime}\in Y^{\prime}} \mathrm{P}(\vec{Y}=\vec{y} \mid Y=y,Y^{\prime}=y^{\prime}, X=x) \mathrm{P}(Y^{\prime}=y^{\prime} \mid Y=y, X=x)  \vec{\mathcal{L}}(\vec{y}, f(x)))\\\nonumber
=& \sum_{{y}=1}^{C} \mathrm{P}(Y=y \mid X=x) \\\nonumber
& (\sum_{\vec{y} \in 2^{[C]}}\sum_{y^{\prime}\in Y^{\prime}} \mathrm{P}(\vec{Y}=\vec{y} \mid Y=y,Y^{\prime}=y^{\prime}, X=x)  \bar{T}_{yy^{\prime}} \vec{\mathcal{L}}(\vec{y}, f(x)))\\\nonumber
=& \sum_{{y}=1}^{C} \mathrm{P}(Y=y \mid X=x) \\\nonumber
\end{align}
\end{minipage}}
\end{equation}
and
\begin{equation}
\begin{minipage}{\linewidth}
\begin{align}
\mathcal{R}(\mathcal{L}, f(X)) &=\mathbb{E}_{Y \mid X}[\mathcal{L}(Y, f(x)) \mid X=x] \nonumber\\\nonumber
&=\sum_{y=1}^{C} \mathcal{L}(y, f(x)) \mathrm{P}(Y=y \mid X=x).\nonumber\\\nonumber
\end{align}
\end{minipage}
\end{equation}
\newpage
Since $\mathrm{P}(\vec{Y}=\vec{y} \mid Y=y,X=x) = 0$ for $\vec{y}$ that does not have $y$, under the condition that 

\begin{equation}
\begin{minipage}{\linewidth}
\begin{align}
&{\mathcal{L}}({y}, f(x))\nonumber\\\nonumber
=& \sum_{{y}=1}^{C} \mathrm{P}(Y=y \mid X=x)\sum_{\vec{y} \in 2^{[C]}}\sum_{y^{\prime}\in Y^{\prime}} \mathrm{P}(\vec{Y}=\vec{y} \mid Y=y,Y^{\prime}=y^{\prime},X=x)  \bar{T}_{yy^{\prime}}\vec{\mathcal{L}}(\vec{y}, f(x))\nonumber\\\nonumber
& = \sum_{\vec{y} \in \vec{\mathcal{Y}} y}  \sum_{{y}=1}^{C} \sum_{y^{\prime}\in Y^{\prime}}  \mathrm{P}(Y=y \mid X=x)\prod_{b^{\prime} \in \vec{y}, b^{\prime} \neq y,  } p_{b^{\prime}} \cdot \prod_{t^{\prime} \notin \vec{y}}\left(1-p_{t^{\prime}}\right)\bar{T}_{yy^{\prime}}\vec{\mathcal{L}}(\vec{y}, f(x))\nonumber\\\nonumber
&=\sum_{\vec{y} \in \vec{\mathcal{Y}} ^{y}} \prod_{b^{\prime} \in \vec{y}, b^{\prime} \neq y,  } p_{b^{\prime}} \cdot \prod_{t^{\prime} \notin \vec{y}}\left(1-p_{t^{\prime}}\right)\vec{\mathcal{L}}(\vec{y}, f(x)).\nonumber\\\nonumber
\end{align}
\end{minipage}
\end{equation}
\subsubsection {The Proof for Lemma 4}
\begin{equation}
\begin{minipage}{\linewidth}
\begin{align}
\mathcal{L}(y, f(x))=\sum_{\vec{y} \in \vec{\mathcal{Y}} ^{y}} \prod_{b^{\prime} \in \vec{y}, b^{\prime} \neq y,  } p_{b^{\prime}} \cdot \prod_{t^{\prime} \notin \vec{y}}\left(1-p_{t^{\prime}}\right)\vec{\mathcal{L}}(\vec{y}, f(x)) =  \vec{\mathcal{L}}(\vec{y}, f(x)),\nonumber\\\nonumber
\end{align}
\end{minipage}
\end{equation}
Ultimately, we can conclude that
\begin{equation}
\begin{minipage}{\linewidth}
\begin{align}
\hat{\mathcal{R}}(\vec{\mathcal{L}}, f(x))=\mathcal{R}(\mathcal{L}, f(x)).\nonumber\\\nonumber
\end{align}
\end{minipage}
\end{equation}
The proof is completed.
\subsubsection{The Proof for Theorem 3}
The goal is to design a new loss function that will enable the hypothesis with adversary-aware partial labels to converge to the optimal classifier trained with true labels. We define $\vec{\mathcal{L}}$ as the new proposed loss function for the adversary-aware partial labels learning. Subsequently, the true and empirical loss function regarding the adversary-aware partial labels is stated as $\hat{R}(f)=\mathbb{E}_{(X, \vec{{Y}}) \sim P_{(X \vec{{Y}}})}[\vec{\mathcal{L}}(f({X}), \vec{{Y}})]$ and $\hat{R}_{pn}(f)=\frac{1}{n} \sum_{i=1}^{n}
\vec{\mathcal{L}}\left(f\left({x}_{i}\right),\vec{{y}}_{i}\right)$, correspondingly. Moreover, we have defined $\left\{\left(\mathbf{x}_{i}, \vec{{y}}_{i}\right)\right\}_{1 \leq i \leq n}$ as the adversary-aware partial label sample space. The functions $\hat{f}^{*}$ and $\hat{f}_{pn}$ are the optimal classifier with minimum expected risk function $\hat{R}(f)$ and empirical $\hat{R}_{pn}(f)$ risk function respectively. Specifically, the model is formalised as $\hat{f}^{*}=\arg\min_{f\in\mathcal{F}}\hat{R}(f)$ and $\hat{f}_{pn}=\arg \min _{f \in\mathcal{F}}\hat{R}_{pn}(f)$. The objective of the newly proposed loss function $\vec{\mathcal{L}}$ is to ensure the convergence of the classifier trained with sample adversary-aware partial label to the optimal classifier trained with population dataset with true labels. Formally, the convergence of $\hat{f}_{pn} \stackrel{n}{\longrightarrow} f^{\star}$ is obtained.
\\
\\
$\textbf{Definition}$. 
Lets denote $\vec{y}_{k}$ as $k$th element of the vector $\vec{y}$ being 1 and others being 0 if $\vec{y}_{k}$ $\in$ $\vec{{y}}$. The $\vec{{y}}$ is a candidate set of the adversary-aware partial label of an instance. Based on Lemma 1 and Theorem 1, the estimation error bound has been proven through
\\
\\
\begin{equation}
\resizebox{1\hsize}{!}{
\begin{minipage}{\linewidth}
\begin{align}
&\hat{R}\left(\hat{f}_{pn}\right)-\min _{f \in F} \hat{R}(f)=\hat{R}\left(\hat{f}_{pn}\right)-\hat{R}\left({\hat{f}}^{\star}\right)\nonumber\\ &=\hat{R}\left(\hat{f}_{pn}\right)-\hat{R}_{pn}(\hat{f})+\hat{R}_{pn}(\hat{f})-\hat{R}_{pn}\left(\hat{f}^{\star}\right)+\hat{R}_{pn}\left(\hat{f}^{\star}\right)-\hat{R}\left(\hat{f}^{\star}\right) \nonumber\\
& \leq \hat{R}\left(\hat{f}_{pn}\right)-\hat{R}_{pn}(\hat{f})+\hat{R}_{pn}\left(\hat{f}^{\star}\right)-\hat{R}\left(\hat{f}^{\star}\right) \nonumber\\
& \leq 2 \sup _{f \in \mathcal{F}}\left|\hat{R}(f)-\hat{R}_{pn}(f)\right| \nonumber\\
& \leq 4 {\Re}\left(\mathcal{F}_{v}\right)+M \sqrt{\frac{\log \frac{2}{\delta}}{2 n}} \nonumber\\
& \leq 4\sqrt{2} L \sum_{k=1}^{c} \Re_{n}\left(\mathcal{F}_{\vec{y}_k}\right)+M \sqrt{\frac{\log \frac{2}{\delta}}{2 n}}.\nonumber\\\nonumber
\end{align}
\end{minipage}}
\end{equation}
Given $\hat{R}_{pn}(\hat{f})-\hat{R}_{pn}\left(f^{\star}\right) \leq 0$, the first inequality equation is established.
The first three equations proof have been shown in \cite{mohri2018foundations}.
\newline
The whole proof is based according to \cite{bartlett2002rademacher}.

\textbf{The definition 1}
Suppose a space $D$ and a sample distribution $D_{S}$ are given in which $S=\left\{s_{1}, \ldots, s_{n}\right\}$ is a set of examples drawn independent, identically distributed from the distribution $D_{S}$. In addition, $\mathcal{F}$ is defined as a class of functions $f: S \rightarrow \mathbb{R}$.
The empirical Rademacher complexity of $\mathcal{F}$ is defined as
\begin{equation}
\resizebox{1\hsize}{!}{
\begin{minipage}{\linewidth}
\begin{align}
\hat{\Re}_{n}(\mathcal{F})=\mathbb{E}_{\sigma}\left[\sup _{f \in \mathcal{F}}\left(\frac{1}{n} \sum_{i=1}^{n} \sigma_{i} f\left(x_{i}\right)\right)\right].\nonumber\\\nonumber
\end{align}
\end{minipage}
}
\end{equation}
The expected Rademacher complexity of the function space 
$\mathcal{F}$ is denoted as 
\begin{equation}
\resizebox{1\hsize}{!}{
\begin{minipage}{\linewidth}
\begin{align}
{\Re}=\mathbb{E}_{D_{S}} \mathrm{E}_{\sigma}\left[\sup _{f \in \mathcal{F}}\left(\frac{1}{n} \sum_{i=1}^{n} \sigma_{i} f\left(x_{i}\right)\right)\right].\nonumber\\\nonumber
\end{align}
\end{minipage}
}
\end{equation}
The independent random variables $\sigma_{1}, \ldots, \sigma_{m}$ are uniformly selected from $\{-1,1\}$. We have defined the random variables as Rademacher variables.
${M}$ is the upper bound of the loss function. Subsequently, for any $\delta>0$, we will have at least probability $1-\delta$
\begin{equation}
\resizebox{1\hsize}{!}{
\begin{minipage}{\linewidth}
\begin{align}
\sup _{f \in \mathcal{F}}\left|\hat{R}(f)-\hat{R}_{pn}(f)\right| \leq 2 \Re(\vec{\mathcal{L}}\circ \mathcal{F})+M \sqrt{\frac{\log 1 / \delta}{2 n}},\nonumber
\end{align}
\end{minipage}
}
\end{equation}
where
\begin{equation}
\resizebox{1\hsize}{!}{
\begin{minipage}{\linewidth}
\begin{align}
\Re(\vec{\mathcal{L}}\circ \mathcal{F})=\mathbb{E}\left[\sup _{f \in \mathcal{F}} \frac{1}{n} \sum_{i=1}^{n} \sigma_{i} \vec{\mathcal{L}}\left(f\left(X_{i}\right), \vec{Y}_{i}\right)\right],\nonumber\\\nonumber
\end{align}
\end{minipage}
}
\end{equation}
is the function space with the expected Rademacher complexity and $\left\{\sigma_{1}, \cdots, \sigma_{n}\right\}$ are Rademacher variables which takes with value of positive and negative 1, such as $\{-1,1\}$ with uniform probability. 
The modified loss function $\vec{\mathcal{L}}$ has been defined in the following equations
\begin{equation}
\resizebox{1\hsize}{!}{
\begin{minipage}{\linewidth}
\begin{align}
\vec{\mathcal{L}}(f(X), \vec{{Y}}) &=-\sum_{i=1}^{c}(\bar{q}_{i}) \log \left(\left((\mathbf{(\bar{T}+I)}^{\top} {f(X))_{i}}\right)\right),\nonumber
\end{align}
\end{minipage}
}
\end{equation}
\begin{equation}
\resizebox{1\hsize}{!}{
\begin{minipage}{\linewidth}
\begin{align}
\mathcal{F}_{\mathrm{V}}=\left\{({X}, \vec{{Y}}) \mapsto \sum_{i=1}^{c} (\bar{q}_{i}) \log \left(\left((\mathbf{(\bar{T}+I)}^{\top} {f(X))_{i}}\right)\right) \mid f \in \mathcal{F}\right\},\nonumber
\end{align}
\end{minipage}
}
\end{equation}
\begin{equation}
\resizebox{1\hsize}{!}{
\begin{minipage}{\linewidth}
\begin{align}
\sup _{f \in \mathcal{F}}\left|\hat{R}(f)-\hat{R}_{pn}(f)\right| \leq 2 \Re(\mathcal{F}_{\mathrm{V}})+M \sqrt{\frac{\log 1 / \delta}{2 n}}.\nonumber\\\nonumber
\end{align}
\end{minipage}
}
\end{equation}
According to McDiarmid's inequality \cite{mcdiarmid1989method}, for any $\delta >0$, with probability at least 1-$\delta/2$ the following equitation holds, namely 
\begin{equation}
\resizebox{1\hsize}{!}{
\begin{minipage}{\linewidth}
\begin{align}
\sup _{f \in \mathcal{F}}\left|\hat{R}(f)-\hat{R}_{pn}(f)\right| \leq \mathbb{E}\left[\sup _{f \in \mathcal{F}}\left|\hat{R}(f)-\hat{R}_{pn}(f)\right|\right] +M \sqrt{\frac{\log 1 / \delta}{2 n}},\nonumber\\\nonumber
\end{align}
\end{minipage}
}
\end{equation}
applying the symmetrization property \cite{vapnik1999nature} that we can acquire the following
\begin{equation}
\resizebox{1\hsize}{!}{
\begin{minipage}{\linewidth}
\begin{align}
\mathbb{E}\left[\sup _{f \in \mathcal{F}}\left|\hat{R}(f)-\hat{R}_{pn}(f)\right|\right] \leq 2 {\mathfrak{R}}\left(\mathcal{F}_{\mathrm{V}}\right).\nonumber\\\nonumber
\end{align}
\end{minipage}
}
\end{equation}
Assume the loss function $\vec{\mathcal{L}}\left(f(\boldsymbol{X}), \vec{{Y}}\right)$ has satisfied the L-Lipschitz property  with respect to $f(\boldsymbol{X})(0<L<\infty)$ with all $\vec{y}_{k} \in \vec{\mathcal{Y}}$ and lastly regarding to the Rademacher vector contraction inequality rule \cite{maurer2016vector} the inequality can be held
\begin{equation}
\resizebox{1\hsize}{!}{
\begin{minipage}{\linewidth}
\begin{align}
{\mathfrak{R}}\left(\mathcal{F}_{V}\right) \leq \sqrt{2} L \sum_{k=1}^{c} \Re_{n}\left(\mathcal{F}_{\vec{y}_{k}}).\right.\nonumber\\\nonumber
\end{align}
\end{minipage}
}
\end{equation}
\\
The proof is completed.
\subsubsection {The Proof for Lemma 4b} 
Since the loss function has been modified, we will show proof of the modified loss function. The modified loss function consisted of two components, the cross entropy loss function $\bar{\mathcal{L}}$ and a transition matrix and identity matrix. In this section, we introduce the modified loss function $\bar{\mathcal{L}}$ and proven through 
\begin{equation}
\resizebox{1\hsize}{!}{
\begin{minipage}{\linewidth}
\begin{align}
\vec{\mathcal{L}}(f(X), \vec{{Y}}) &=-\sum_{i=1}^{c}(\bar{q}_{i}) \log \left(\left((\mathbf{(\bar{T}+I)}^{\top} {f(X))_{i}}\right)\right), \nonumber\\
&=-\sum_{i=1}^{c} \boldsymbol{1}(\bar{q}_{i}) \log \left(\frac{\sum_{j=1}^{c} (\bar{T}_{j i})) \exp \left(g_{j}(X)\right)}{\sum_{k=1}^{c} \exp \left(g_{k}(X)\right)}\right),\nonumber\\\nonumber
\end{align}
\end{minipage}
}
\end{equation}
in which $(\mathbf{(T+I)}^{\top}{f(X)})_{i}$ is defined as the $i$-th row of $(\mathbf{T+I})^{\top}{f}; {h}: \mathcal{X} \rightarrow \mathbb{R}^{c}, f_{i}(X) \in \mathcal{H}, \forall i \in[c];$
In addition $f_{i}(X)=\frac{\exp \left(g_{i}(X)\right)}{\sum_{k=1}^{c} \exp \left(g_{k}(X)\right)}.$
\newline
The proof is completed.
\subsubsection{Additional Experimental Details}
We have compared with most recent partial label learning algorithms, which are PICO \cite{wang2022pico}, LWS\cite{pmlr-v139-wen21a}, and PRODEN\cite{pmlr-v119-lv20a} on CIFAR-10\cite{krizhevsky2009learning}, CIFAR-100\cite{krizhevsky2009learning} and CUB200\cite{WahCUB_200_2011}.
The negative and rival labels of adversary-aware partial labels datasets are generated according to the probability $q_{b,l}^{*}: =\mathcal{P}({b,l}\in{{\vec{Y}}} \mid Y=y,Y^{\prime}=l, X=x)$ with $b$ $\neq$ $y$. The class instance-dependent partial labels are manually generated. We have used the $\pm0.02$ proportion of the output $\Delta(f_{i}(X))$ corresponding to each instance after the softmax layer from the pre-trained classifier Resent18 \cite{He_2016_CVPR}. More specifically, we have defined all $C-1$ negative label where $\bar{y} \neq y$ with a uniform probability to be flipped to false positive. Finally, the probability can be defined as $q^{*}_{b,l}\pm{0.02}$.
The projection head of the contrastive network has 128-dimensional embedding with a 2-layer MLP. The data augmentation modules are following the previous work \cite{wang2022pico}. The queue size is fixed at $8192$, $8192$ and $4192$ for the CIFAR-10, CIFAR-100 and CUB200 correspondingly. The momentum coefficients are  $0.999$ for the contrastive network update. The $\alpha$ is the hyperparameter of the immature teacher within momentum (ITWM), controlling the proportion of prototype updates. The $\alpha = $ 0.1 and $\beta$ = $0.01$ are selected for the immature teacher within momentum (ITWM) without adversary-aware loss and the immature teacher within momentum (ITWM). The optimizer SGD with a momentum of 0.9 and 256 batch size are used to train the model for 299 epochs with a cosine learning rate schedule. Except for the total epochs, others are identical to the previous work \cite{wang2022pico}. For the temperature parameter $\tau$, we have set it to $0.07$. The loss weighting factors are set to $\lambda = \{ 0.5 \}$.  The partial label rate at  $q \in \{0.1,0.3,0.5\}$ have been implemented for CIFAR-10 and  $q \in \{0.03,0.05,0.1\}$ for CIFAR-100 and CUB200.

The adversary partial label rate at  $q^{*} \in \{0.1\pm0.02,0.3\pm0.02,0.5\pm0.02\}$ have been implemented for CIFAR-10 and $q^{*}\in \{0.03\pm0.02,0.05\pm0.02,0.1\pm0.02\}$ for CIFAR-100 and CUB200.
Training without contrastive learning for CIFAR-10 is 1 epoch for all the partial rates with respect to clean partial labels. For CIFAR10 adversary-aware partial labels, the setting of 50 epochs training without contrastive learning is applied for  $q$  =$\{0.1,0.3,0.5\}$. We have trained without contrastive learning for the clean partial label with  $q$  = $\{ 0.01,0.05,0.1 \}$ for epochs of $\{ 20,20,100 \}$ on CIFAR-100 and CUB200. Moreover, the epochs of $\{ 20,100,100 \}$ is set for the adversary-aware partial rate at $q^{*}$=$\{0.03\pm0.02,0.05\pm0.02,0.1\pm0.02\}$ of adversary-aware partial labels learning problem on CIFAR-100 and CUB200.
\subsubsection{Additional Experiment for CIFAR-10}
We have verified our method on an additional synthetic dataset, CIFAR-10. The implementation setting is mainly identical to \cite{wang2022pico}. For CIFAR-10 clean partial label learning, we have implemented the experiments according to each baseline's implementation details, and the best results were replicated from the baseline works\cite{wang2022pico}. The CIFAR-10 adversary-aware partial label problem has used the ResNet18 neural network\cite{he2016deep} as the backbone. The $\alpha=$ $0.1$ and $\beta$ = $0.01$ are chosen for the immature teacher within momentum (ITWM) without $\title{T}$ (Clean partial label) and immature teacher within momentum (ITWM) method. The learning rate is $0.01$, and the weight decay is $1e-3$. The $\mathcal{R}\text{esNet}-18$ is used for training. For the clean partial label,  $q$  at $\{ 0.1,0.3,0.5 \}$ is used for the experiments. The adversary-aware partial label is set to  $q^{*}$  = \{ 0.1$\pm{0.02}$, 0.3$\pm{0.02}$, 0.5$\pm{0.02}$ \} for experiments. We have trained the model without contrastive loss for the epochs of $\{1,1,1\}$ with the clean partial label at partial rate of  $q$  = $\{ 0.1,0.3,0.5 \}$. We have trained the model without contrastive learning for epochs of $\{50,50,50\}$ for the adversary-aware partial labels with partial rates of  $q^{*}$  = \{ 0.1$\pm{0.02}$, 0.3$\pm{0.02}$, 0.5$\pm{0.02}$ \}.  
\subsubsection{The Classification Accuracy Comparisons}
Our proposed methods have consistently outperformed the previous works for the most challenging scenarios  $q$  =\{0.5,0.5,0.1\} on CIFAR-10, CIFAR-100 and CUB200.
\begin{figure}[H]
\centering
\subcaptionbox{\label{fig3:a}}{\includegraphics[width=0.35\textwidth]{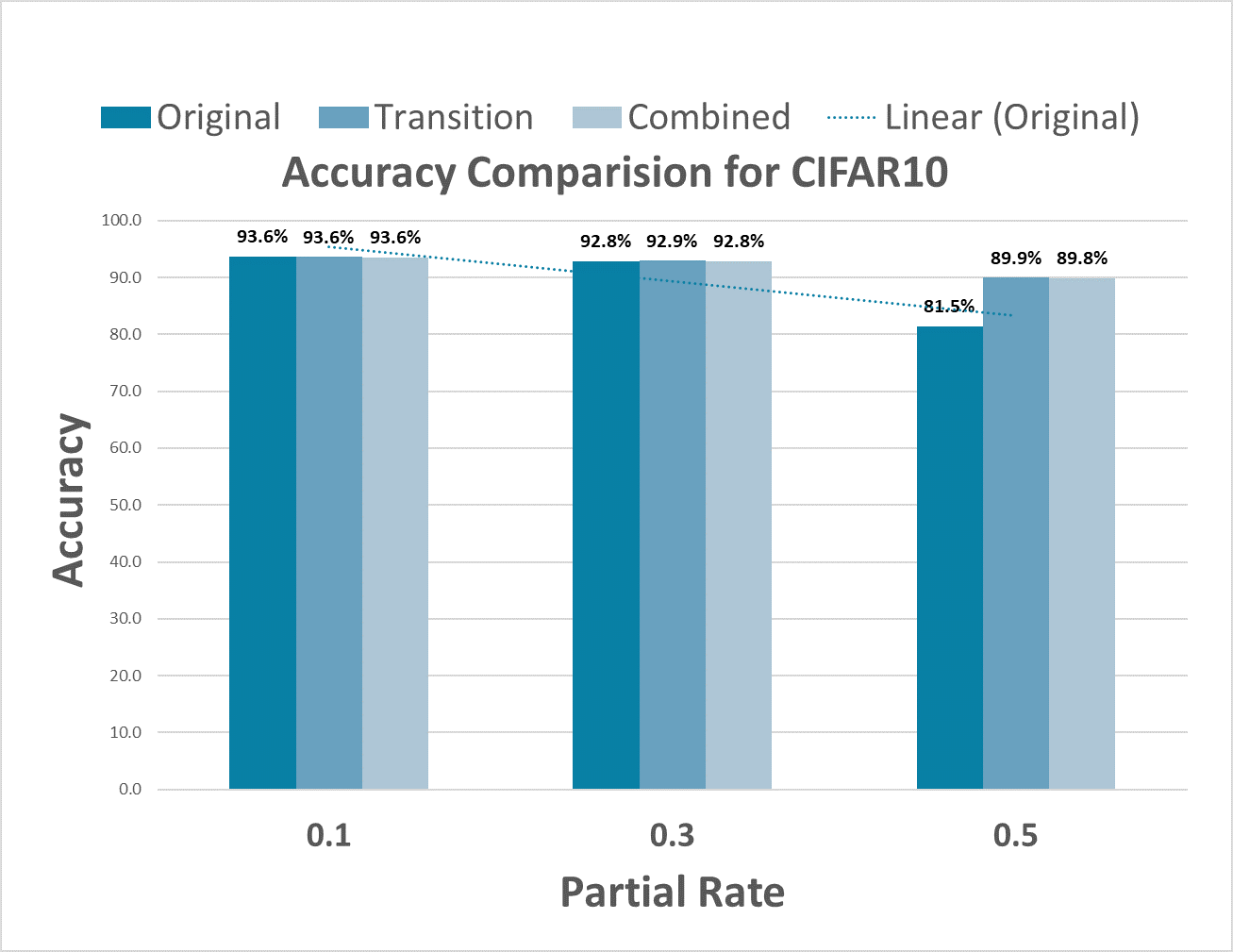}}\hspace{1em}%
\subcaptionbox{\label{fig3:b}}{\includegraphics[width=0.35\textwidth]{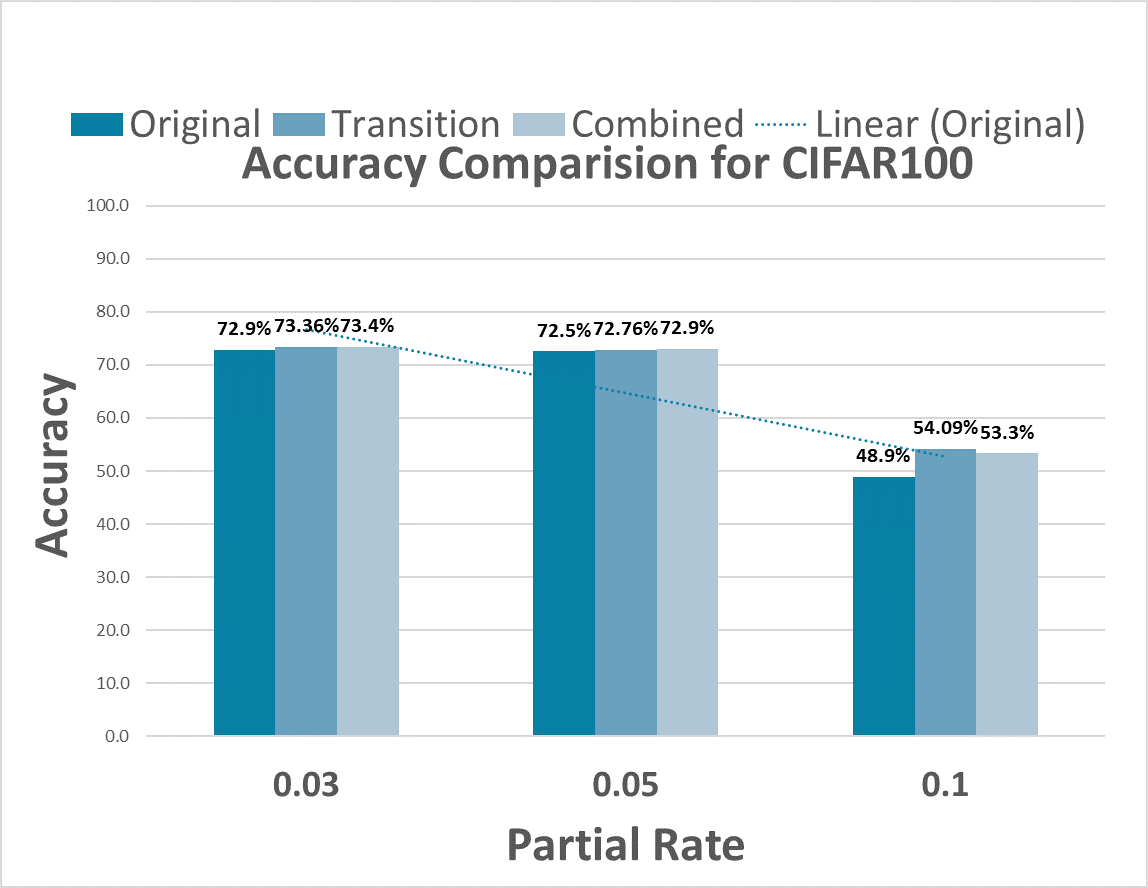}}
\subcaptionbox{\label{fig3:c}}{\includegraphics[width=0.35\textwidth]{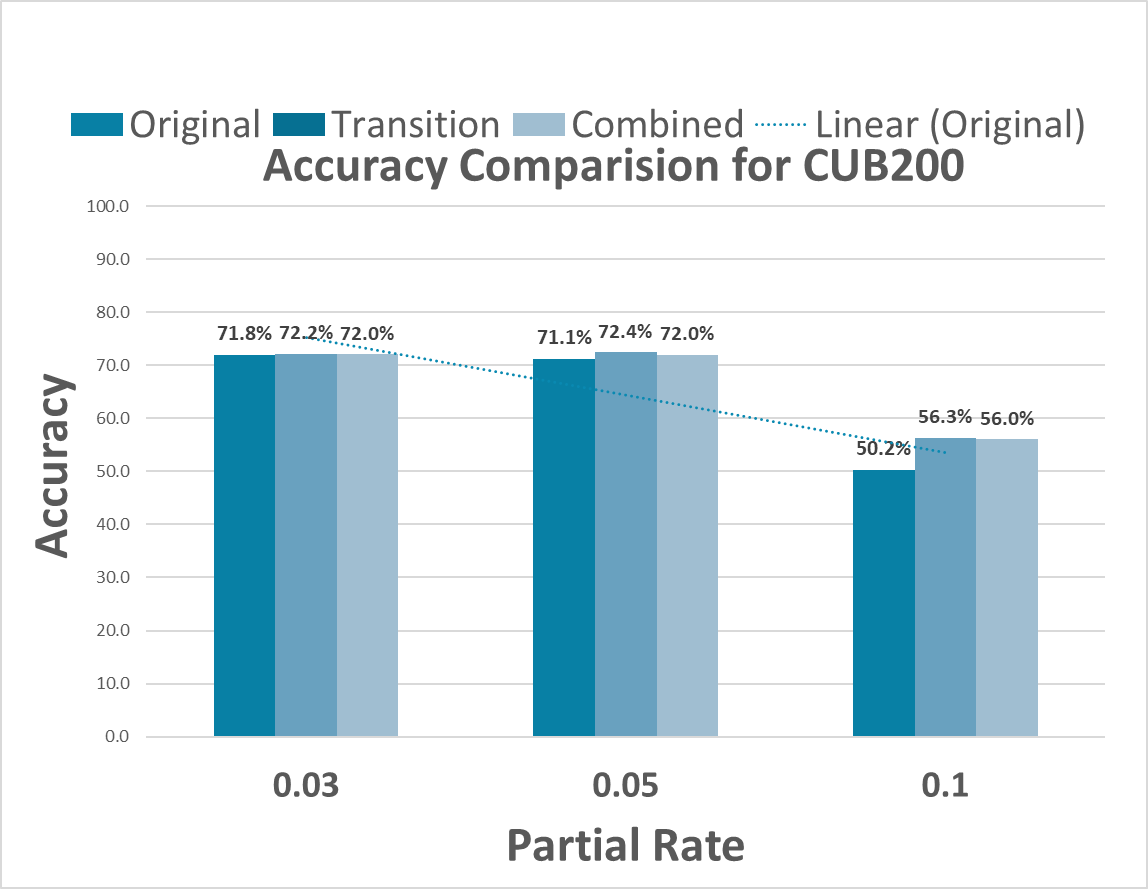}}
\caption{The Classification Accuracy Comparisons} 
\end{figure}
\subsubsection{The Hyperparameter Comparisons}
We have also conducted a comparative analysis on the impact of hyperparameter $\alpha$ on the final classification performance. The larger the hyperparameter, the better the classification performance. Our proposed method has compared the hyperparameter $\alpha$ at \{0.1,0.5,0.9\} for all dataset. The $\alpha=$ 0.1 has been chosen throughout the experiments. 
\begin{figure}[H]
\centering
\captionsetup{font=small}
\subcaptionbox{\label{fig1:a}}{\includegraphics[width=0.4\textwidth]{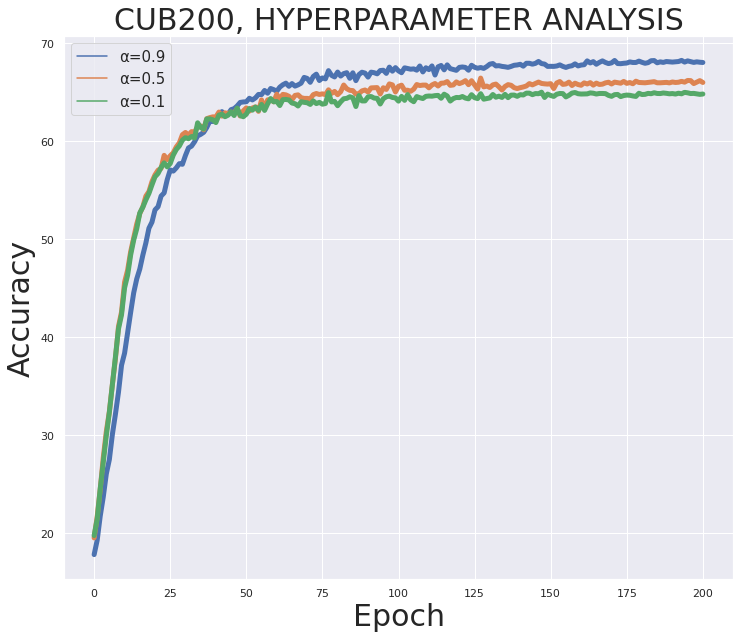}}\hspace{1em}%
\caption{The Classification Accuracy of our proposed method using $\alpha$ =\textbf{[0.1,0.5,0.9]} for CUB200 } 
\label{est_error}
\end{figure}
\subsubsection{Adversary-Aware Loss Comparison.} 
Figure 3 shows the experimental result comparisons for CIFAR100 between the modified loss function and cross-entropy loss function before and after the momentum updating strategy. Our method achieves SOTA performance. The adversary-aware matrix plays an indispensable role. In the first stage, the divergence becomes more apparent as the epoch reaches 100 epochs for CIFAR100 in Top-1 classification accuracy. The comparison demonstrated that the modified loss function works consistently throughout the whole stage of learning, especially for the more challenging learning scenario where the partial rate is at 0.1. 
\begin{figure}[H]
\centering
\captionsetup{font=small}
\label{est_error}
\centering
\subcaptionbox{\label{fig4:a}}{\includegraphics[width=0.4\textwidth]{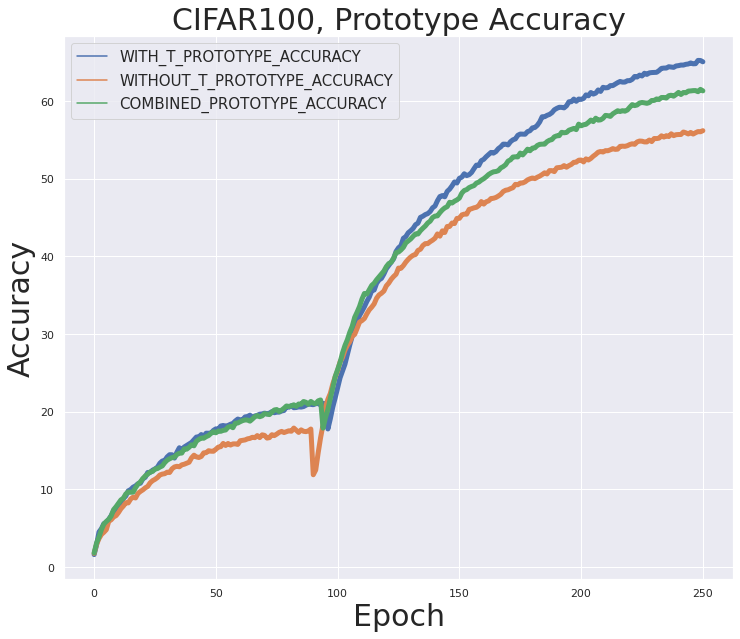}}\hspace{1em}%
\subcaptionbox{\label{fig4:a}}{\includegraphics[width=0.4\textwidth]{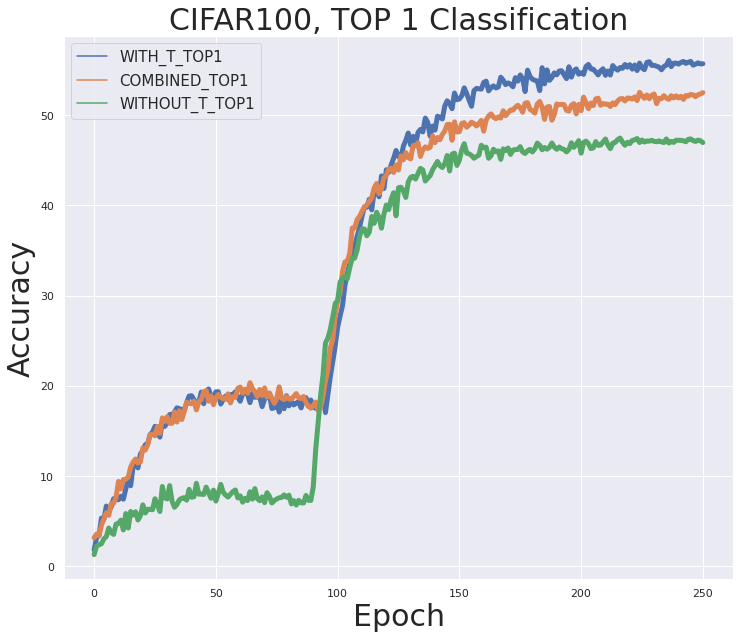}}\hspace{1em}%
\caption{The Top 1 and Prototype Accuracy of the Proposed Method and the Method in \cite{wang2022pico} PiCO on CIFAR100.}
\label{est_error}
\end{figure}
\subsection{Implementation Details}
\textbf{Adversary-Aware Matrix.}
The transition matrix is a common tool for building statistically consistent classifiers in noise label and complementary label problems \cite{yu2018learning,huang2006correcting,liu2014learnability}. In this paper that we have introduced the adversary-aware matrix for building statistically consistent classifiers for the adversary-aware partial label problem. The Adversary-Aware Matrix $ T \in \mathcal{R}^{c \times c}$ is constructed as $T_{y,y^{\prime}}=\bar{T}+I$.  We set the diagonal element of $\tilde{T}$ to one to ensure $y \in \vec{y}$.
\newline
\textbf{New rival Label.}
The rival label is generated according to the label noise transition matrix $\bar{T}$. We have defined ordinary partial label generation $B$. The $B$ is defined as $\mathrm{P}(\vec{Y} \mid X) $, general partial label generation, is defined accordingly as E.q 2. In application, we can randomly give out proportional of survey with rival and other without the rival according to the adversary aware matrix. This will ensure that adversary will not be able to retrieve the insightful data by purposely enquiry a participant to reveal given out answers.
By formulating the rival as $\boldsymbol{R}=\mathrm{P}(\vec{Y}\mid X)$, which equal to
$\displaystyle\min\{1,B^{(2^{c}-2) \times c }\bar{T}^{c,c}\}$
and $R_{i,j} \in[0,1]^{(2^{c}-2) \times c}$, for $\forall_{i,j} \in[c]$. We now have the adversary aware partial label.

\subsection{Ablation Study for \texorpdfstring{$\bar{T}$}{T-bar}}
In the following, we have shown how the classification performance is impacted if the entries of the class instance-dependent transition matrix $\bar{T}_{\textbf{Original}}$ is updated to $\bar{T}_{\textbf{New}}$ to show the robustness of our proposed method. For instance if the number of class is equal to 10, then the entries of $\bar{T}$ is defined as below. In our problem setting, each row has five entries equal to 0.2 in the original $\bar{T}$ and  has five entries equal to 0.3 each row for new  $\bar{T}$. 
\begin{align*}
\bar{T}_{\textbf{Original}}=
\begin{bmatrix}
0    & 0.2  & 0  & 0.2  & 0.2  & 0.2\\
0.2  & 0    & 0.2  & 0.2  & 0  & 0.2\\
0.2  & 0.2  & 0    & 0.2  & 0  & 0.2\\ 
0.2  & 0.2  & 0  & 0    & 0.2  & 0.2\\ 
0.2  & 0.2  & 0  &0.2   & 0    & 0.2\\
0.2  & 0  & 0.2  & 0.2  & 0.2  & 0 \\
\end{bmatrix}
\bar{T}_{\textbf{New }}=
\begin{bmatrix}
0    & 0.3  & 0  & 0.3  & 0.3  & 0.3\\
0.3  & 0    & 0.3  & 0.3  & 0  & 0.3\\
0.3  & 0.3  & 0    & 0.3  & 0  & 0.3\\ 
0.3  & 0.3  & 0  & 0    & 0.3  & 0.3\\ 
0.3  & 0.3  & 0  &0.3   & 0    & 0.3\\
0.3  & 0  & 0.3  & 0.3  & 0.3  & 0 
\end{bmatrix}
\end{align*}
\begin{center}
\begin{tabular}{ c|c|c } 
\hline
Data &Method& $q^{*}$=0.1\\ 
\textbf{CIFAR100} & PiCO\cite{wang2022pico} & 20.941(24.015)$\%$\\
\hline
Data &Method& $q^{*}$=0.1\\ 
\textbf{CIFAR100} & ATM & \textbf{54.156(0.066)$\%$}\\ \hline
\end{tabular}
\end{center}

\begin{center}
\begin{tabular}{ c|c|c } 
\hline
Data &Method& $q^{*}$=0.1\\ 
\textbf{CUB200} & PiCO\cite{wang2022pico} &21.22(-25.155)$\%$\\
\hline
Data &Method& $q^{*}$=0.1\\ 
\textbf{CUB200} & ATM & \textbf{48.62(-7.64)}$\%$\\ \hline
\end{tabular}
\end{center}
\subsection{Why adversary aware partial label learning is a more challenging problem} By adding the rival, the partial label generation process compare with the new label generation is easier:
\\
\\
\begin{align*}
A=
\begin{bmatrix}
0    & 0.25  & 0  & 0.25  & 0.25  & 0.25\\
0.2  & 0    & 0.25  & 0.25  & 0  & 0.25\\
0.2  & 0.25  & 0    & 0.25  & 0  & 0.25\\ 
0.25  & 0.25  & 0  & 0    & 0.25  & 0.25\\ 
0.25  & 0.25  & 0  &0.25   & 0    & 0.25\\
0.25  & 0  & 0.25  & 0.25  & 0.25  & 0 
\end{bmatrix}
\end{align*}
\begin{align*}
B=
\begin{bmatrix}
1    & 0.5\pm{0.02}  & 0.5\pm{0.02}  & 0.5\pm{0.02}  & 0.5\pm{0.02}  & 0.5\pm{0.02}\\
0.5\pm{0.02}  & 1    & 0.5\pm{0.02}  & 0.5\pm{0.02}  & 0  & 0.5\pm{0.02}\\
0.5\pm{0.02}  & 0.5\pm{0.02}  & 1    & 0.5\pm{0.02}  & 0  & 0.5\pm{0.02}\\ 
0.5\pm{0.02}  & 0.5\pm{0.02}  & 0  & 1    & 0.5\pm{0.02}  & 0.5\pm{0.02}\\ 
0.5\pm{0.02}  & 0.5\pm{0.02}  & 0  &0.5\pm{0.02}   & 1    & 0.5\pm{0.02}\\
0.5\pm{0.02}  & 0  & 0.5\pm{0.02}  & 0.5\pm{0.02}  & 0.5\pm{0.02}  & 1 
\end{bmatrix}
\end{align*}
By adding the rival using the label noise transition matrix as such $A \times B$ and we can conclude that $[A \times B]_{ij}>[B]_{ij}$. Even though noises added to the partial label noise has made the problem more challenging, unless the adversary-aware transition matrix is given, it will greatly help us to reduce the uncertainty from the transition matrix.
\subsubsection{Algorithm Table}
\begin{algorithm}[H]
  \scriptsize
  \caption{\footnotesize Adversary Aware Partial label learning}\label{euclid}
  \textbf{Goal:} Minimise the Total loss function $\lambda$ 
  \textbf{Input:} The Adversary-Aware PLL $\mathcal{\bar{D}}$ and Batch size Samples $\mathcal{\bar{D}}_{b}.$
  \textbf{Output:} The optimal $W$ of the Total Loss Function.
  \begin{algorithmic}[H]
    \For{\texttt{$\textbf{x_{i}} \in \textbf{Total Epochs}$}}
    \State$ \vec{D}_{b} \in \vec{D}$
    \State$ D_{q}=\{\boldsymbol{u}_{i}= f\left(\operatorname{Aug}_{q}\left(\boldsymbol{x}_{i}\right)\right) \mid \boldsymbol{x}_{i} \in \vec{D_{b}}\}$
    \State$D_{k}=\{\boldsymbol{z}_{i}=f^{\prime}\left(\operatorname{Aug}_{k}\left(\boldsymbol{x}_{i}\right)\right) \mid \boldsymbol{x}_{i} \in \vec{D_{b}}\}$
    \State$\bar{\mathcal{C}}=D_{q} \cup D_{k} \cup \text { queue }$
      \For{\texttt{$\textbf{x}_{i} \in \vec{D}$}}
      \State$\hat{y}_{i}=\arg \max _{c \in Y_{i}} f^{c}\left(\operatorname{Aug}_{q}\left(\boldsymbol{x}_{i}\right)\right)$
     \State$\boldsymbol{v}_i^{t+1} = \sqrt{1-\alpha^2} \boldsymbol{v}_i^{t} +  \alpha  \frac{\boldsymbol{g}}{\| \boldsymbol{g}\|_2}$
     \State $N_{+}{\left(\boldsymbol{x}_{i}\right)}=\left\{\boldsymbol{z}^{\prime} \mid \boldsymbol{z}^{\prime} \in \bar{\mathcal{C}}\left(\boldsymbol{x}_{i}\right), \bar{y}^{\prime}=(\hat{y}_{i}=c)\right\})$ 
      \EndFor
      \For{\texttt{$\textbf{u}_{i} \in D_{q}$}}
      \State $\quad r_{c}= \begin{cases}1 & \text { if } c=\arg \max _{j \in Y} \boldsymbol{u_{i}}^{\top} \boldsymbol{v}_{j} \\ 0 & \text { otherwise }\end{cases},$
        \State $\boldsymbol{\bar{q}}=\phi \boldsymbol{\bar{q}}+(1-\phi)  \boldsymbol{r_{c}}$

      \EndFor
        \State \texttt{$\mathcal{L}$=$\lambda\mathcal{L}_(f(x_{i}), \tau, C)$+$\vec{\mathcal{L}}(f(x_{i}), \vec{Y})$}\Comment{Equation 18 + Equation 17}
        \State \texttt{$\mathcal{L}$=$\lambda\mathcal{L}+ \vec{\mathcal{L}}$}\Comment{Total Loss}
    \EndFor
  \end{algorithmic}
\end{algorithm}
\end{document}